\documentclass[10pt,twocolumn,letterpaper]{article}

\usepackage[pagenumbers]{wacv}              %

\usepackage{graphicx}
\usepackage{comment}
\usepackage{amsmath}
\usepackage{amssymb}
\usepackage{booktabs}
\usepackage{diagbox} %
\usepackage{tabularray}
\usepackage{multirow}
\usepackage{boldline}
\usepackage[dvipsnames]{xcolor}

\usepackage[pagebackref,breaklinks,colorlinks]{hyperref}

\definecolor{LinkColor}{HTML}{105FC1}
\hypersetup{
    colorlinks=true,   
    linkcolor=black,
    filecolor=black, 
    urlcolor=LinkColor,
    pdftitle={RGB-X Object Detection via Scene-Specific Fusion Modules},
}

\usepackage[capitalize]{cleveref}
\crefname{section}{Sec.}{Secs.}
\Crefname{section}{Section}{Sections}
\Crefname{table}{Table}{Tables}
\crefname{table}{Tab.}{Tabs.}

\def\wacvPaperID{681} %
\def\confName{WACV}
\def\confYear{2024}

\begin{document}

\title{RGB-X Object Detection via Scene-Specific Fusion Modules}

\author{Sri Aditya Deevi$^{1}$\footnotemark[1] \quad\quad\quad
Connor Lee$^{1}$\footnotemark[1] \quad\quad\quad
Lu Gan$^{1}$\footnotemark[1] \quad\quad\quad
Sushruth Nagesh$^{2}$ \\
Gaurav Pandey$^{2}$ \quad\quad\quad Soon-Jo Chung$^{1}$ \medskip \\ 
$^{1}$California Institute of Technology \quad $^{2}$Ford Motor Company \\
\texttt{\small \{sdeevi, clee, ganlu, sjchung\}@caltech.edu} \quad
\texttt{\small \{snagesh1, gpandey2\}@ford.com}
}

\maketitle
{
    \renewcommand{\thefootnote}{\fnsymbol{footnote}}
    \footnotetext[2]{This work was funded by the Ford University Research Program.}
    \footnotetext[1]{Equal contribution}
}

\begin{abstract}
    Multimodal deep sensor fusion has the potential to enable autonomous vehicles to visually understand their surrounding environments in all weather conditions. However, existing deep sensor fusion methods usually employ convoluted architectures with intermingled multimodal features, requiring large coregistered multimodal datasets for training. In this work, we present an efficient and modular RGB-X fusion network that can leverage and fuse pretrained single-modal models via scene-specific fusion modules, thereby enabling joint input-adaptive network architectures to be created using small, coregistered multimodal datasets. Our experiments demonstrate the superiority of our method compared to existing works on RGB-thermal and RGB-gated datasets, performing fusion using only a small amount of additional parameters. Our code is available at~\href{https://github.com/dsriaditya999/RGBXFusion}{https://github.com/dsriaditya999/RGBXFusion}.
\end{abstract}

\section{Introduction}
\label{sec:intro}
Autonomous vehicles rely on object detection algorithms to understand and interact with their surrounding environments. In order to be robust against different driving conditions, these algorithms operate on data from various sensor modalities ranging from optical cameras to LiDAR, each with their own advantages and disadvantages. Because no single sensor modality is robust to all possible conditions that may be encountered during driving, multiple sensor modalities are often used in conjunction via \textit{deep sensor fusion (DSF)} to boost performance during normal driving operations, as well as to ensure segmentation and object detection reliability in adverse weather conditions~\cite{feng}.   

\begin{figure}
    \centering
    \includegraphics[width=\columnwidth]{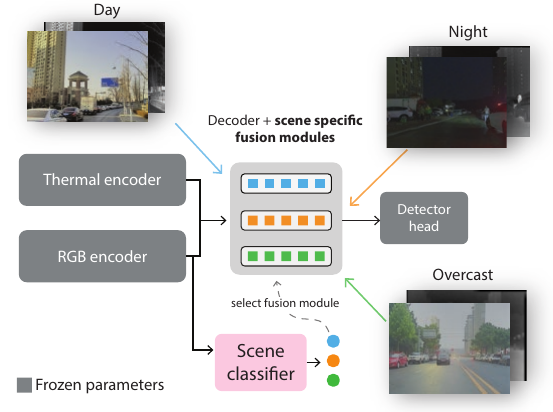}
    \caption{Our multimodal object detection approach combines RGB and thermal pretrained networks using lightweight, scene-specific fusion modules. Fusion modules are trained using categorized scene images and are used adaptively during inference with an auxiliary scene classifier.}
    \label{fig:high-level-arch}
\end{figure}

Unlike traditional sensor fusion which merges processed sensor data outputs coming from independent pipelines, current works in DSF generally require joint end-to-end training of multi-branch sensor networks on large multimodal datasets~\cite{Bijelic_2020_STF, cao2023multimodal, zhang2020multispectral, liang2023explicit, qingyun2021cross, zhang2021guided} such as NuScenes~\cite{nuscenes2019}, Berkeley Deep Drive~\cite{bdd100k}, and Waymo~\cite{sun2020scalability} prior to deployment in the wild~\cite{feng}.
This means that fusion architectures must undergo time-consuming and potentially expensive retraining (in cost and carbon emissions) anytime a sensor modality is removed or added~\cite{patterson2021carbon}, and that they fail to take full advantage of state-of-the-art RGB pretrained networks.

In this paper, we propose the use of existing, well-known attention blocks as lightweight, scene-specific attention modules in order to easily fuse pretrained networks and to better adapt to common weather disturbances. We demonstrate our approach (\cref{fig:high-level-arch}) for object detection applications, training RGB-thermal and RGB-gated fusion models on RGB, thermal, and gated imagery collected in adverse driving conditions such as night, fog, snow, and rain~\cite{Bijelic_2020_STF, liu2022target, flir}. We also leverage the attention modules as a method to visually interpret the contributions of each sensor modality. Compared to prior works, our approach takes us another step closer to enabling a modular, \textit{drag-and-drop} design for deep sensor fusion that absolves the need for extensive and expensive retraining while delivering on-par or better performance. \textbf{Our contributions are as follows:}
\textbf{1.} A lightweight, modular RGB-X fusion network for object detection that leverages pretrained single-modality networks. \textbf{2.} A scene-adaptive fusion approach that selectively uses different fusion modules for different scene/weather conditions. \textbf{3.} Extensive experiments on publicly available RGB-X datasets that demonstrate the superiority of our approach in terms of detection performance and computational efficiency.

\section{Related Work}

\textbf{Object Detection:} %
Most modern methods for detecting objects utilize convolutional neural networks (CNN) or transformers. CNN object detectors include two-stage and single-shot detectors~\cite{ren2015faster, redmon2018yolov3, liu2016ssd, tan2020efficientdet}. A two-stage detector has an additional region proposal step while a single-shot detector relies only on a feature extractor and a detection head that directly predicts bounding boxes and classes, resulting in faster inference~\cite{zou2023object}. To deploy on mobile devices, neural architecture search (NAS) has been used to develop faster and lighter networks and detection architectures~\cite{howard2019searching, tan2020efficientdet}. In this work, we adopt the EfficientDet~\cite{tan2020efficientdet} detection architecture to target self-driving car applications that operate on mobile computing devices. Recent large vision transformer models have achieved state-of-the-art object detection results, but are not suitable for real-time use on robotic platforms~\cite{carion2020end, liu2021swin}. 

\textbf{Deep Sensor Fusion:}
Robotic perception applications, notably for self-driving cars, rely on DSF to add sensor redundancy and to increase perception robustness and performance in both common and adverse operating scenarios. Current DSF algorithms consume multimodal data using deep networks and are trained end-to-end, combining different features at various points throughout a network depending on their particular fusion policy~\cite{feng, fayyad_deep_2020}. Early fusion policies aggregate raw inputs or features extracted early on in the network~\cite{wagner_multispectral_2016, liu_multispectral_2016} while mid-fusion approaches~\cite{liang2018deep, 8578131} operate on deeper, intermediate representations. Late fusion methods operate directly on bounding box outputs and can be used directly with pretrained detectors, but are subject to the performance of pretrained models~\cite{chen2022multimodal}. In our work, we opt for a mid-fusion approach in order to take full advantage of the different feature modalities at various stages. 

Regardless of fusion policies, current DSF algorithms and datasets for self-driving cars mainly focus on incorporating sensors like LiDAR and radar with RGB cameras~\cite{8578131, bdd100k, nuscenes2019, sun2020scalability, geiger2013vision}. In our work, we are interested in supplementing RGB with 2D image data from thermal and gated cameras due to the rich semantic information they provide and their robustness to fog and lighting in driving scenarios~\cite{feng}.  

\textbf{RGB-Thermal Object Detection:} Current RGB-thermal (RGB-T) object detection methods typically operate on aligned RGB-thermal image pairs and utilize some form of attention-based modules to perform mid-fusion on RGB and thermal image features. \cite{zhang2021guided} utilizes intra-modality and inter-modality spatial attention modules to enhance and adaptively fuse intermediate features, respectively, prior to passing downstream to a detection head. Recently, \cite{cao2023multimodal} proposed mid-fusion modules that utilize channel attention to dynamically swap RGB and thermal feature channels. This helps to maximize feature usefulness before enhancing local features via parameter-free spatial attention. Other works including \cite{fu2023lraf, zhu2023multi, qingyun2021cross} fuse multi-modal data in a similar fashion but instead leverage transformer-based attention modules that increase model size and computational cost. \cite{Bijelic_2020_STF} does not use thermal images, but similarly fuses RGB, gated, and projected LiDAR and radar data using local entropy masks in lieu of attention. In our work, we demonstrate that pretrained, single-modality detectors can be fused using simple, scene-specific channel and spatial attention modules to achieve strong RGB-T object detection performance.

\begin{figure*}
    \centering
    \includegraphics[width=\linewidth]{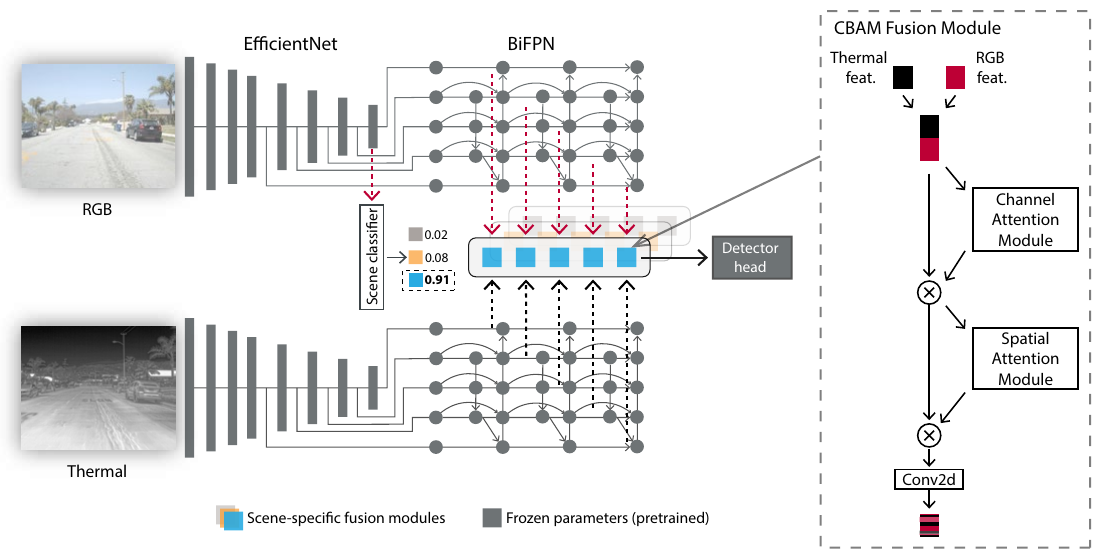}
    \caption{Overall framework of our scene-adaptive CBAM model for RGB-X fusion illustrated by RGB-T fusion. RGB and thermal images are processed by separate EfficientNet backbones, followed by BiFPNs. The features from BiFPNs are used for cross-modal feature fusion using modules selected by the scene classifier. The detector head utilizes these fused features to obtain the final detection results. The right side of the figure illustrates the CBAM fusion module, consisting of channel and spatial attention blocks, for feature fusion. }
    \label{fig:arch}
\end{figure*}

\section{Approach}

We propose a modular RGB-X fusion network for object detection that is built upon pretrained single-modal detection architecture and multi-stage convolutional block attention modules (CBAM)~\cite{woo2018cbam} for cross-modal feature fusion. This modularity separates the training of single-modal backbones that contain the majority of network parameters and the training of a small fusion module, mitigating the requirement of large-scale multi-modal training data. The overall architecture for RGB-X fusion is shown in \cref{fig:arch} using RGB-T as an example. We have an individual EfficientDet~\cite{tan2020efficientdet} for each image modality consisting of an EfficientNet~\cite{tan2019efficientnet} backbone network, a bidirectional feature pyramid network (BiFPN) and a detector head. While we choose to use EfficientDet to demonstrate our approach, we note that this architecture can be built using any single-modal detection network.

We employ CBAM to fuse the RGB and thermal features output from the respective BiFPN at various stages. Each CBAM fuses features at the same scale, resulting in 5 CBAM fusion modules. During training, only CBAM parameters are updated while pretrained object detector weights are frozen. CBAM modules are trained per scene category and are selected for use during inference time using an auxiliary scene classifier. 
In the rest of this section, we go over the details of our fusion mechanism and the auxiliary scene classifier, before describing the overall scene-adaptive fusion algorithm for RGB-X object detection. 

\subsection{Convolutional Block Attention Fusion}
We use CBAM to fuse RGB and thermal (or gated) CNN feature maps $\mathbf{F_{rgb}}$ and $\mathbf{F_{x}}$, respectively. We concatenate features from both modalities across the channel dimension to create an input feature map $\mathbf{F}$ for CBAM:
\begin{equation}
    \mathbf{F} = [\mathbf{F_{rgb}}; \mathbf{F_{x}}] \in \mathbf{R}^{B \times C \times H \times W},
\end{equation}
where $B$ denotes the batch size, and $C, H, W$ denote the channel and spatial dimensions of the feature, respectively.
Following the notation in~\cite{woo2018cbam}, a CBAM module takes the feature map $\mathbf{F}$ and masks it using channel and spatial attention operators $M_c$, $M_s$ such that
\begin{align}
    \mathbf{F'} &= M_c(\mathbf{F}) \otimes \mathbf{F}, \\
    \mathbf{F''} &= M_s(\mathbf{F'}) \otimes \mathbf{F'}, 
\end{align}
where $\otimes$ denotes element-wise multiplication.
We further convolve $\mathbf{F''}$ with $C / 2$ kernels resulting in $C/2$ channels which is the original feature dimension.  

Channel attention operator $M_c$ is computed via
\begin{equation}
    M_c(\mathbf{F}) = \sigma(\mathbf{W_1} \mathbf{W_0} \mathbf{F^c_{avg}} + \mathbf{W_1} \mathbf{W_0} \mathbf{F^c_{max}}),
\end{equation}
where $\sigma$, $\mathbf{W}$, $\mathbf{F^c_{avg}}$, $\mathbf{F^c_{max}}$ denotes the sigmoid function, linear layer weights, the global average and max pooled features, respectively. 
Spatial attention is computed via
\begin{equation}
    M_s(\mathbf{F}) = \sigma(f^{7 \times 7}([\mathbf{F^c_{avg}}; \mathbf{F^c_{max}}])),
\end{equation}
where $\mathbf{F^c_{avg}}, \mathbf{F^c_{max}}$ are computed via channel-wise mean and max operations and $f^{7 \times 7}$ denotes convolution with a kernel size of 7.

\subsection{Auxiliary Scene Classification}

We utilize a simple scene classifier during inference time to adaptively select the most suitable set of fusion modules for the current setting, based on the intuition that the fusion module should attend different modalities under different scene/weather conditions. The scene classifier consists of a 2D adaptive average pooling operator followed by a fully connected layer, taking in the features from the RGB object detector encoder and outputs probabilities of possible scene categories. We choose RGB features as the input for scene classification due to their high variance in different scenes. 

\subsection{Scene-Specific Fusion}

We train different CBAM fusion modules for various scenes by considering scene-specific dataset splits (\cref{tab:data-splits}). The number of parameters in different parts of the proposed fusion model is shown in \cref{tab:model_stat}. The total number of trainable parameters per scene is significantly less than the total number of parameters, making our approach expeditious. During inference of scene-adaptive fusion, we use the CBAM fusion modules trained on the scene with the highest probability, as indicated by the scene classifier.

\begin{table}
\centering
\caption{Parameter statistics of the proposed RGB-X fusion model.}
\resizebox{0.8\linewidth}{!}{%
\begin{tabular}{l|c} 
\hlineB{3}
Network Part & \# Parameters \\
\hline
Backbones (RGB + X)              & 24.8 M \\
BiFPNs (RGB + X)                 & 0.12 M  \\
Detection Head   & 1.60 M \\
Fusion Modules (one per fusion level)   & 0.21 M \\
\hline
Total                                   & 26.7 M \\
Total Trainable (per scene)                        & 0.21 M \\
\hlineB{3}
\end{tabular}
}
\label{tab:model_stat}
\end{table}

\section{Results}

\subsection{Implementation and Training Details}
Our code is written in PyTorch and based on the EfficientDet\footnote{\href{https://github.com/rwightman/efficientdet-pytorch}{https://github.com/rwightman/efficientdet-pytorch}} repository. Pretrained RGB detectors on COCO dataset~\cite{coco} were taken from the same repository. All other networks were trained using the Adam optimizer, a batch size of 8, an initial learning rate of 1e$^{-3}$ with an exponential learning rate schedule, and a $L_2$ weight decay of 1e$^{-3}$. The maximum number of epochs is set to $300$ and $50$ for pretraining single modality networks and fine-tuning RGB-X fusion networks, respectively. The scene classifier is trained for $50$ epochs while the RGB backbone remains frozen. Networks were trained using an Nvidia P100 GPU. 

\subsection{Datasets}
We use the following RGB-X datasets to validate our method and compare against state-of-the-art baselines. The train/val/test split statistics we use for various datasets and scene conditions are shown in \cref{tab:data-splits}. 
\begin{table}
    \caption{Dataset scene and train/val/test splits in our experiments.}
    \begin{subtable}[h]{\columnwidth}
        \centering
        \resizebox{0.75\linewidth}{!}{
            \begin{tabular}{c|c|c|c|c|c|c}
                \hlineB{3}
                \multirow{3}{*}{Split} & \multicolumn{6}{c}{Scene Condition} \\ \cline{2-7}
                & \multicolumn{2}{c|}{Clear} & \multicolumn{2}{c|}{Fog}  & \multicolumn{2}{c}{Snow}  \\ \cline{2-7}
                & Day & Night & Day & Night &  Day & Night  \\
                \hline
                Train & 2147 & 1572 & 712 & 438  & 1365 & 1455  \\
                Val & 537 & 393 & 438 & 110 & 342 & 364  \\
                Test & 895 & 655 & 297 & 183 & 570 & 607   \\
                \hlineB{3}
           \end{tabular}
       }
       \caption{Seeing Through Fog}
       \label{tab:stf-splits}
    \end{subtable}
    \hfill
    \begin{subtable}[h]{0.35\columnwidth}
        \centering
        \resizebox{\linewidth}{!}{
            \begin{tabular}{c|c|c}
                \hlineB{3}
                \multirow{2}{*}{Split} & \multicolumn{2}{c}{Scene Condition} \\ \cline{2-3}
                & Day & Night \\
                \hline
                Train & 3476 & 653 \\
                Val & --- & --- \\
                Test & 702 & 311 \\
                \hlineB{3}
           \end{tabular}
       }
       \caption{FLIR Aligned~\cite{zhang2020multispectral}}
       \label{tab:flir-splits}
    \end{subtable}
    \hfill
    \begin{subtable}[h]{0.6\columnwidth}
        \centering
        \resizebox{\linewidth}{!}{
            \begin{tabular}{c|c|c|c|c}
                \hlineB{3}
                \multirow{2}{*}{Split} & \multicolumn{4}{c}{Scene Condition} \\ \cline{2-5}
                & Day & Night & Overcast & Challenge \\
                \hline
                Train & 992 & 488 & 746 & 484 \\
                Val & 216 & 108 & 190 & 122 \\
                Test & 323 & 140 & 205 & 156 \\
                \hlineB{3}
           \end{tabular}
       }
       \caption{M$^3$FD~\cite{liu2022target}}
       \label{tab:m3fd-splits}
    \end{subtable}
     
     \label{tab:data-splits}
\end{table}

\textbf{FLIR Aligned:} The FLIR Aligned dataset~\cite{zhang2020multispectral} consists of 5,142 aligned RGB-thermal image pairs from the original FLIR ADAS object detection dataset~\cite{flir}. This derived dataset consists of bounding box annotations for \textit{person}, \textit{bicycle} and \textit{car} classes. The provided train and test splits contain 4,129 and 1,013 image pairs, respectively. We manually divided them into \emph{day} and \emph{night} scene categories based on the appearance.

\textbf{M$^3$FD:} The M$^3$FD object detection dataset consists of 4,200 coregistered, time-synchronized RGB-thermal image pairs~\cite{liu2022target}. Bounding box annotations for \textit{people, car, bus, motorcycle, truck} and \textit{lamp} classes are provided. The data is also split into four scene categories \textit{(day, night, overcast, challenge)} in~\cite{liu2022target} according to environment characteristics. We use the train/val/test splits provided by~\cite{liang2023explicit} due to the unavailability in the original dataset.

\textbf{Seeing Through Fog:} The Seeing Through Fog (STF) multispectral object detection dataset~\cite{Bijelic_2020_STF} consists of synchronized RGB/gated/LiDAR/radar/unaligned thermal data for a variety of weather conditions. The dataset also provides bounding box annotations for \textit{pedestrian}, \textit{truck}, \textit{car}, \textit{cyclist}, and \textit{dontcare} classes. For training our scene-adaptive model, we considered the scene splits in \cref{tab:stf-splits} due to overlaps in original splits. For evaluation, we follow the original scene splits including \textit{clear}, \textit{light fog}, \textit{dense fog}, and \textit{snow/rain}. We use pairs of aligned 12-bit RGB and 10-bit gated images throughout this work.

\begin{figure}
    \centering
    \begin{subfigure}{0.43\columnwidth}
    \includegraphics[width=0.5\columnwidth]{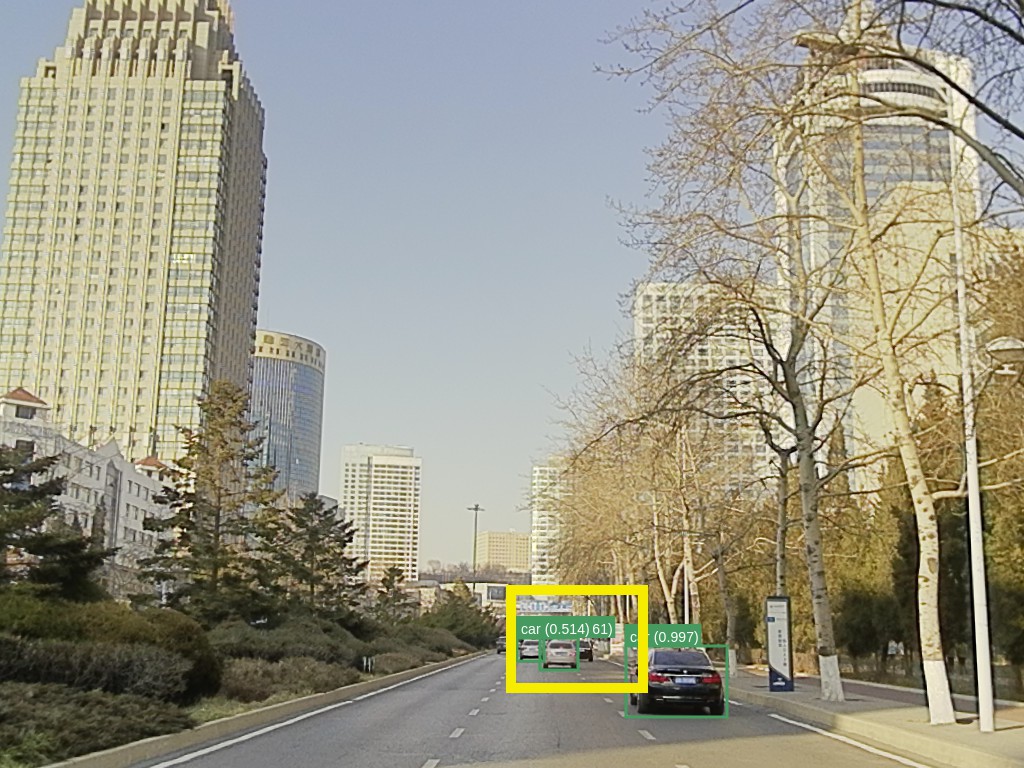}\includegraphics[width=0.5\columnwidth]{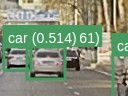}
    \includegraphics[width=0.5\columnwidth]{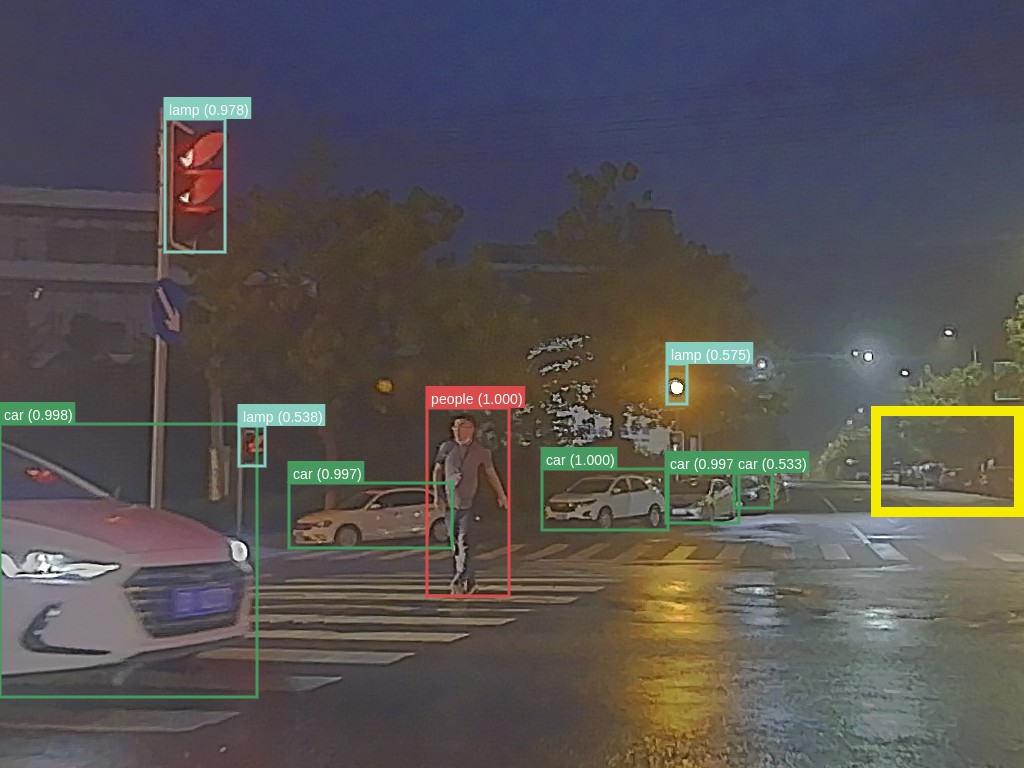}\includegraphics[width=0.5\columnwidth]{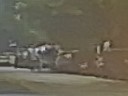}
    \includegraphics[width=0.5\columnwidth]{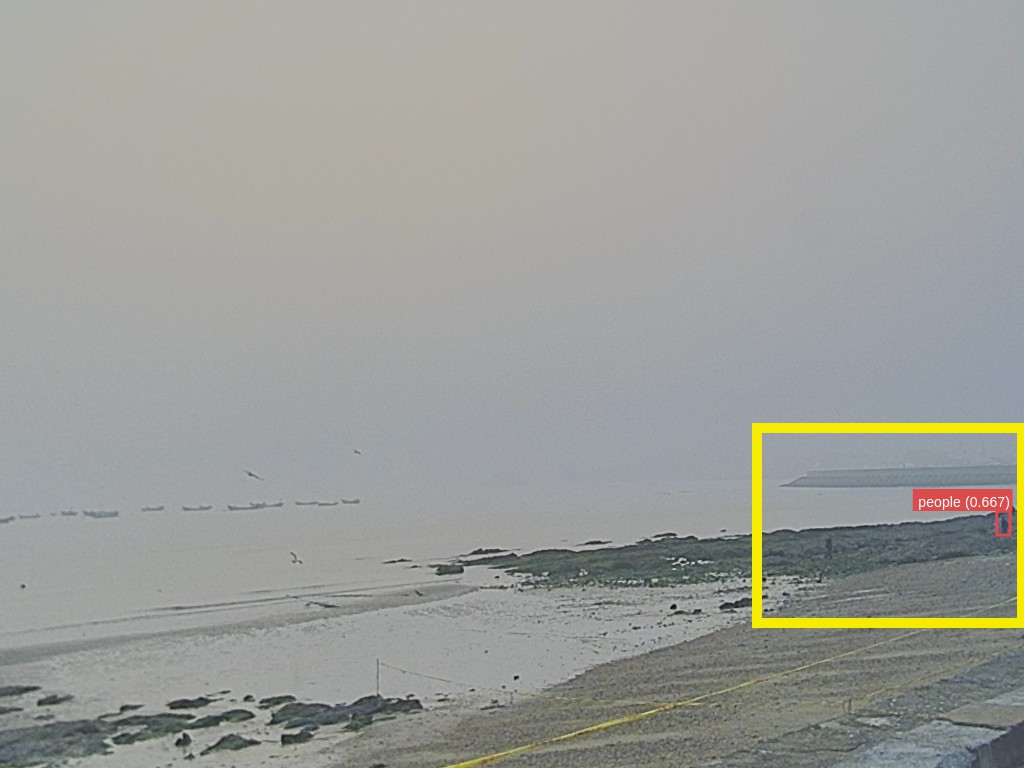}\includegraphics[width=0.5\columnwidth]{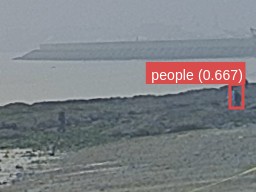}
    \caption{\footnotesize Scene-Agnostic CBAM}
    \end{subfigure}
    \begin{subfigure}{0.43\columnwidth}
    \includegraphics[width=0.5\columnwidth]{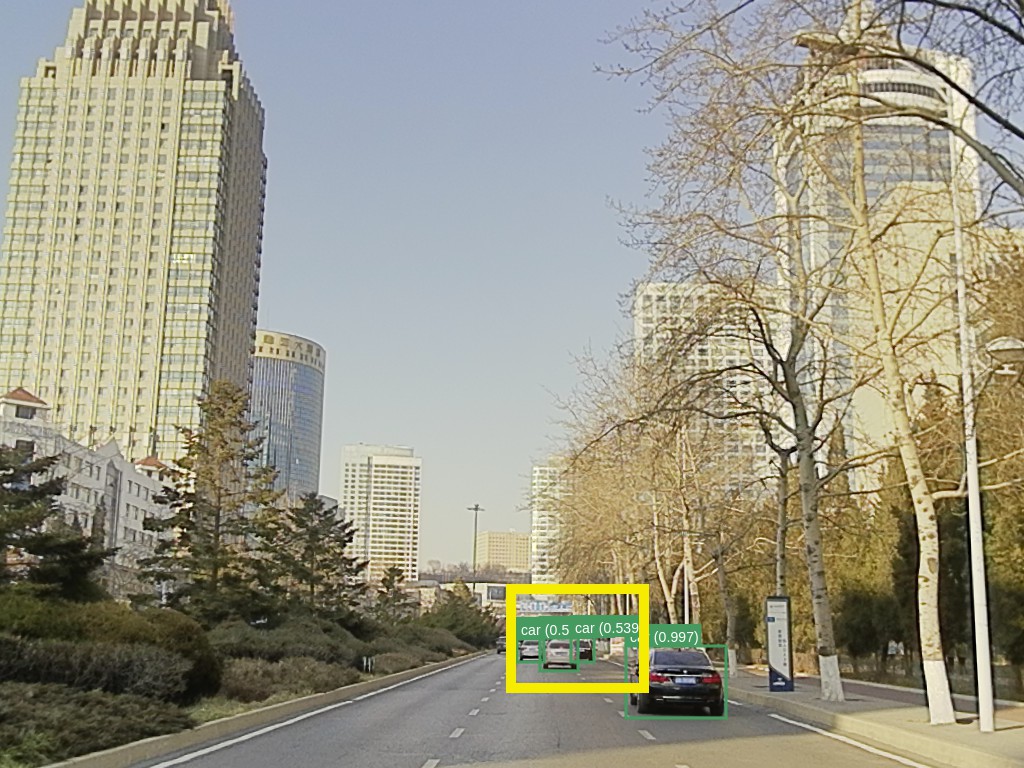}\includegraphics[width=0.5\columnwidth]{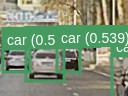}
    \includegraphics[width=0.5\columnwidth]{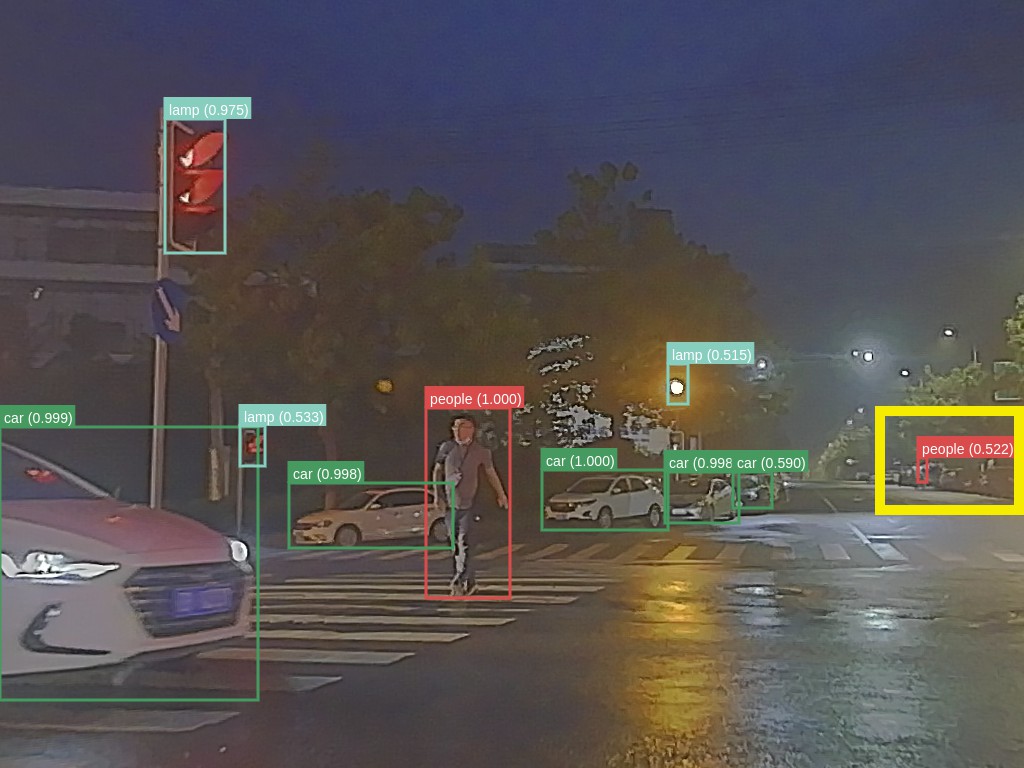}\includegraphics[width=0.5\columnwidth]{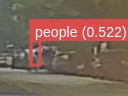}
    \includegraphics[width=0.5\columnwidth]{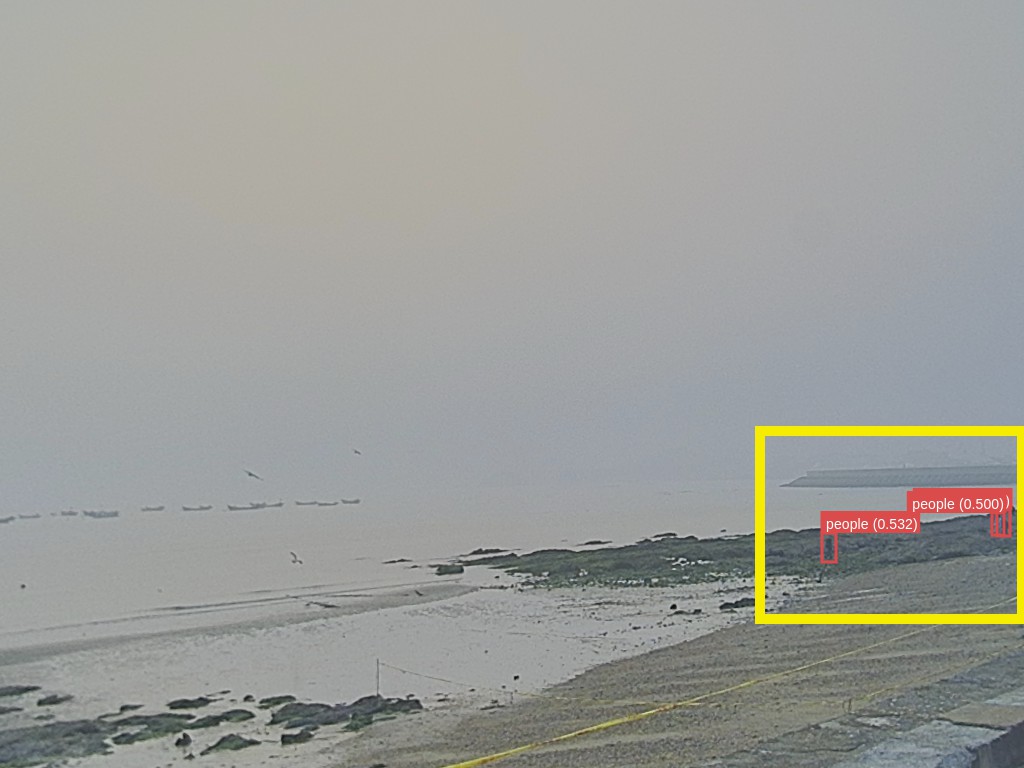}\includegraphics[width=0.5\columnwidth]{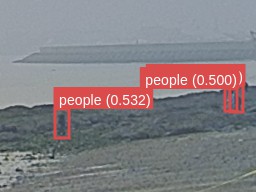}
    \caption{\footnotesize Scene-Adaptive CBAM}
    \end{subfigure}
    
    \caption{Qualitative detection results on M$^3$FD dataset. Zoomed-in images (yellow rectangle) are shown on the right of the original images for better visualization.}
    \label{fig:results-m3fd}
\end{figure}

\begin{figure*}
    \centering
    \begin{subfigure}{0.3\columnwidth}
    \includegraphics[width=1\columnwidth]{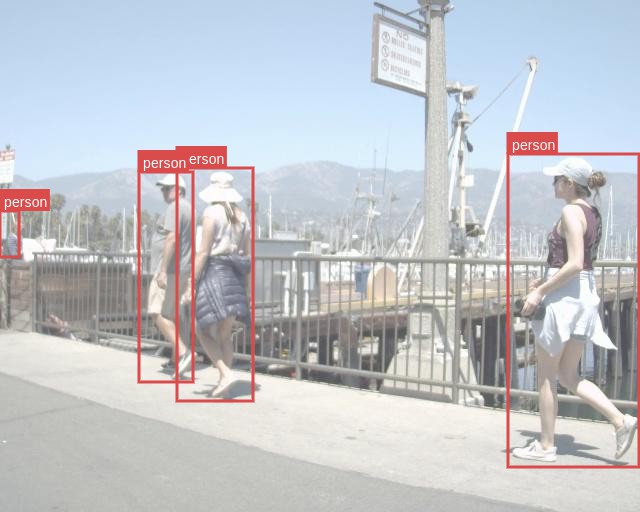}
    \includegraphics[width=1\columnwidth]{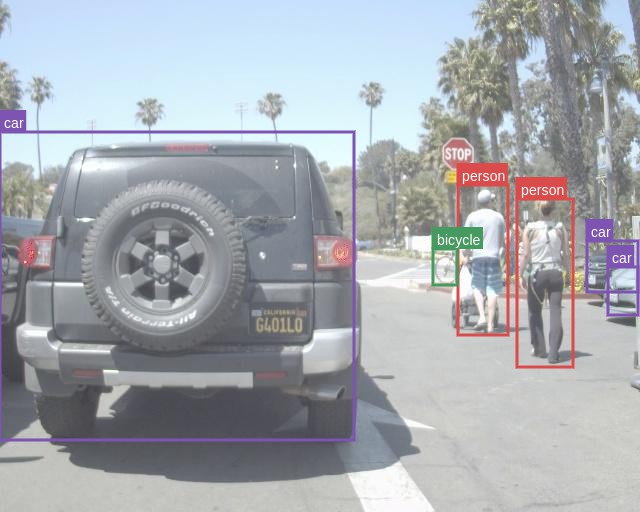}
    \includegraphics[width=1\columnwidth]{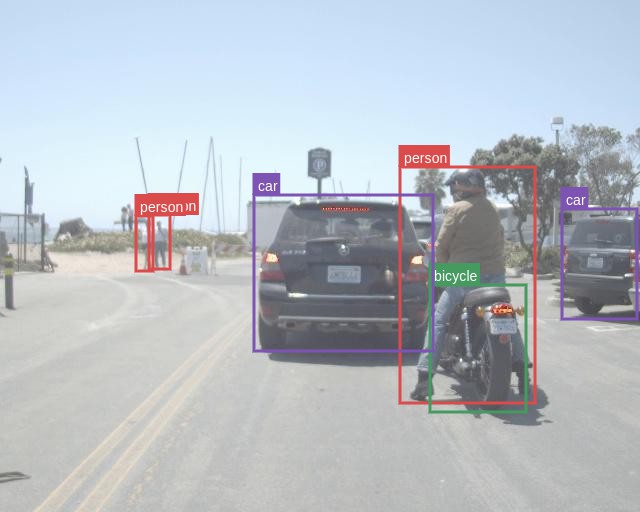}
    \includegraphics[width=1\columnwidth]{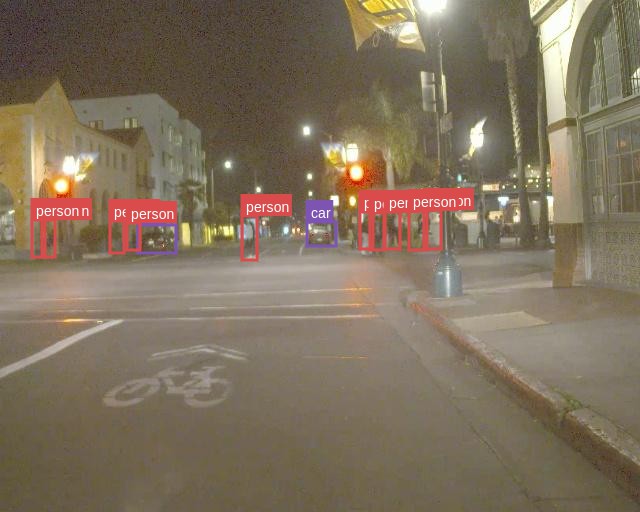}
    \includegraphics[width=1\columnwidth]{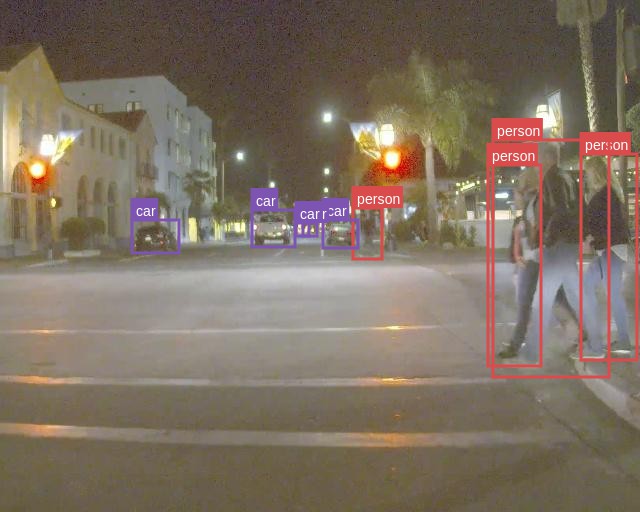}
    \includegraphics[width=1\columnwidth]{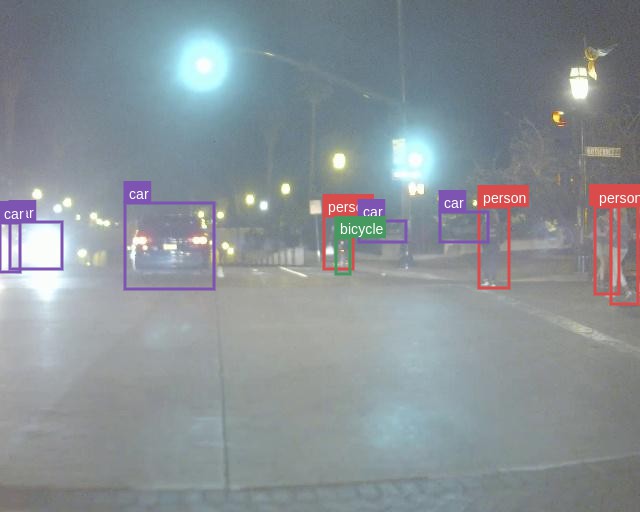}
    \caption{\footnotesize RGB with GT}
    \end{subfigure}
    \begin{subfigure}{0.3\columnwidth}
    \includegraphics[width=1\columnwidth]{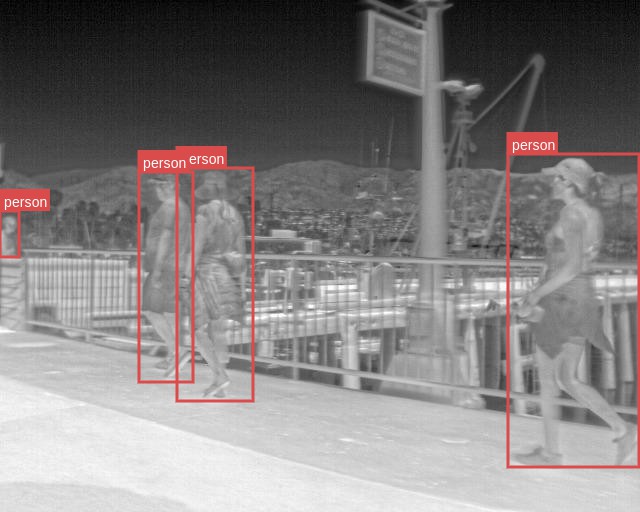}
    \includegraphics[width=1\columnwidth]{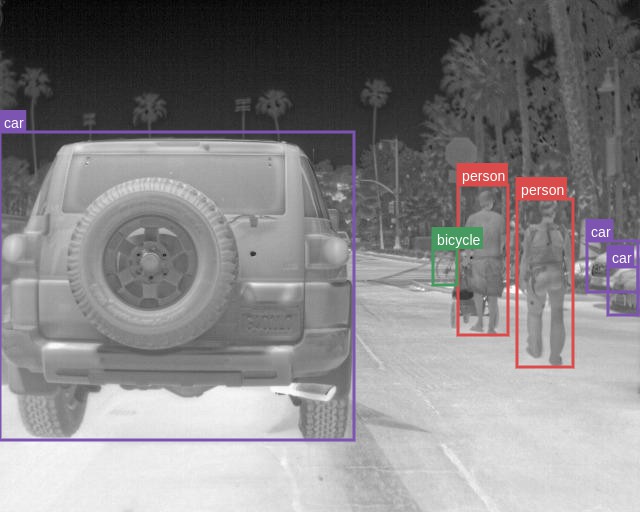}
    \includegraphics[width=1\columnwidth]{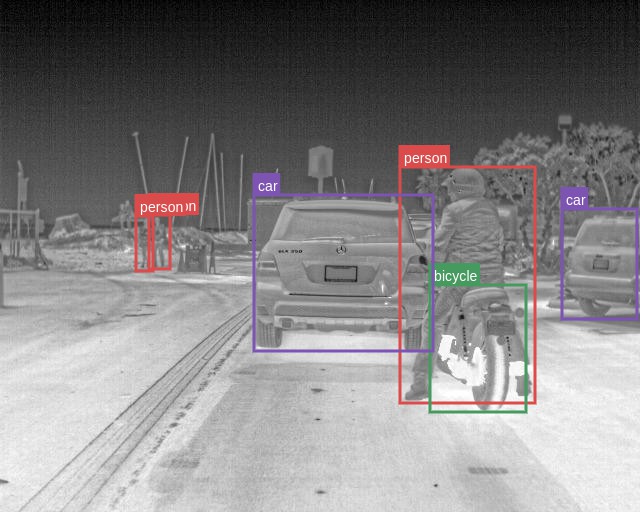}
    \includegraphics[width=1\columnwidth]{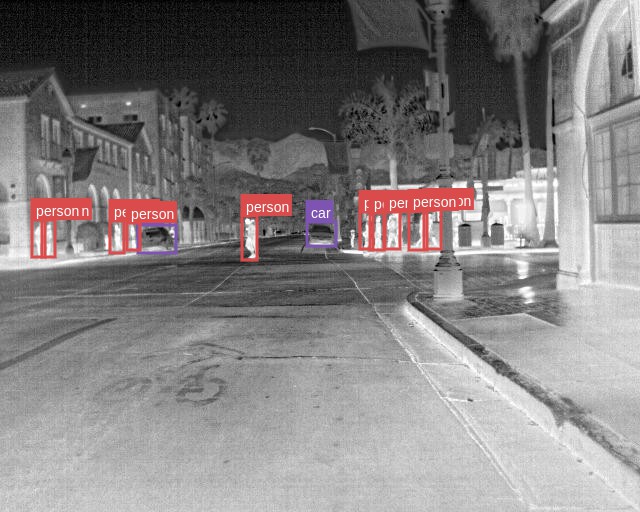}
    \includegraphics[width=1\columnwidth]{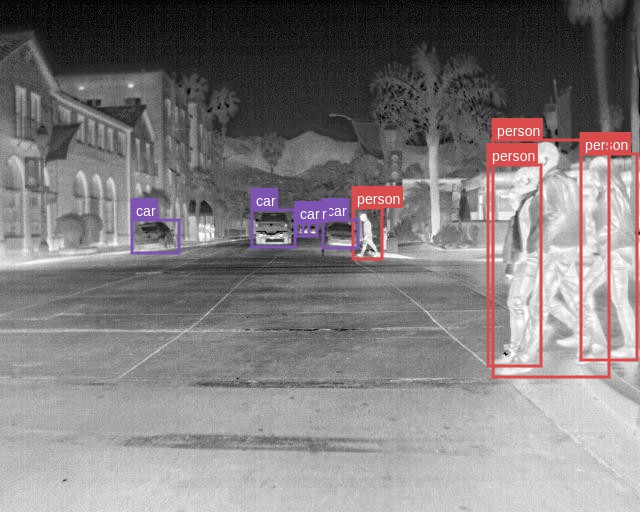}
    \includegraphics[width=1\columnwidth]{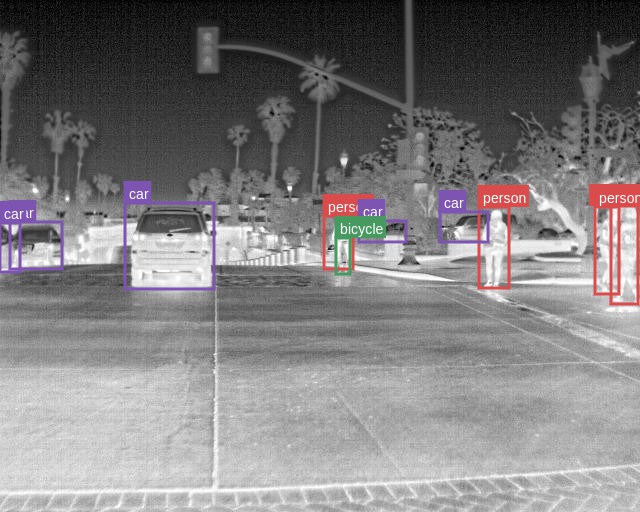}
    \caption{\footnotesize Thermal with GT}
    \end{subfigure}
    \begin{subfigure}{0.3\columnwidth}
    \includegraphics[width=1\columnwidth]{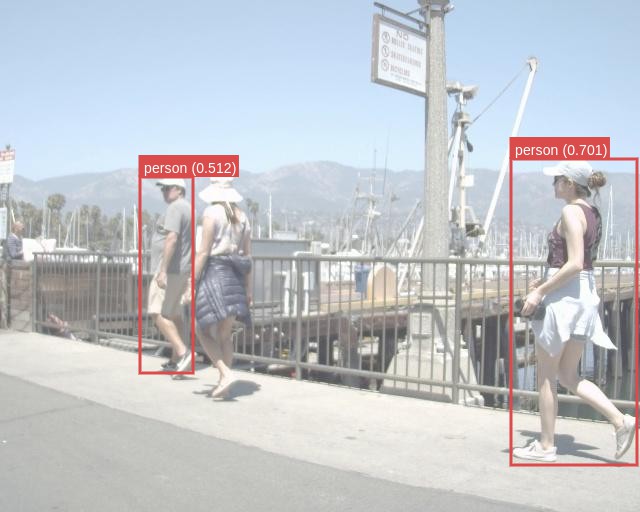}
    \includegraphics[width=1\columnwidth]{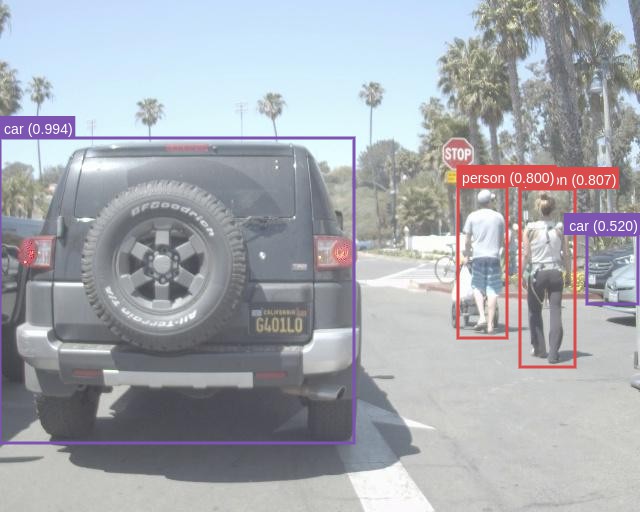}
    \includegraphics[width=1\columnwidth]{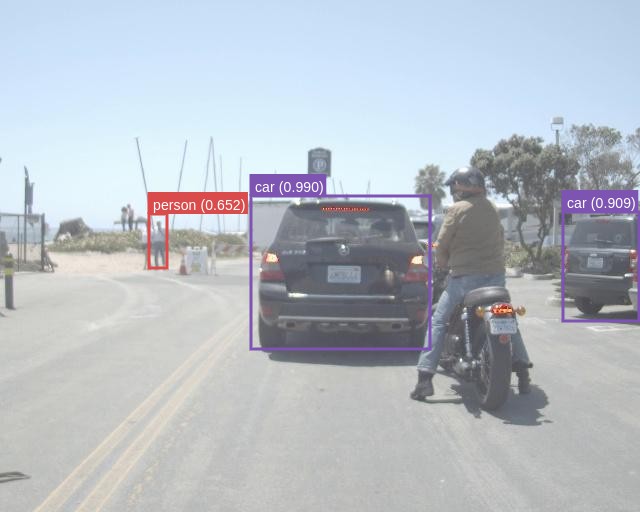}
    \includegraphics[width=1\columnwidth]{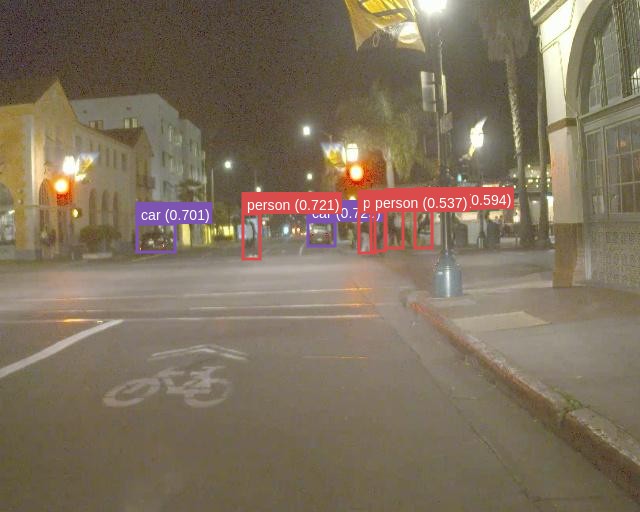}
    \includegraphics[width=1\columnwidth]{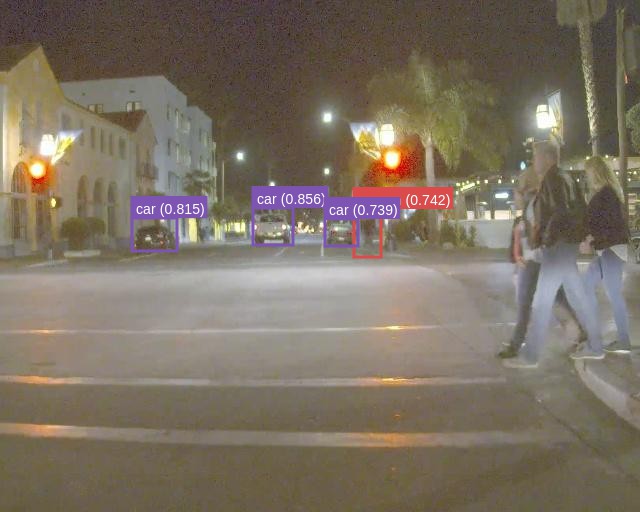}
    \includegraphics[width=1\columnwidth]{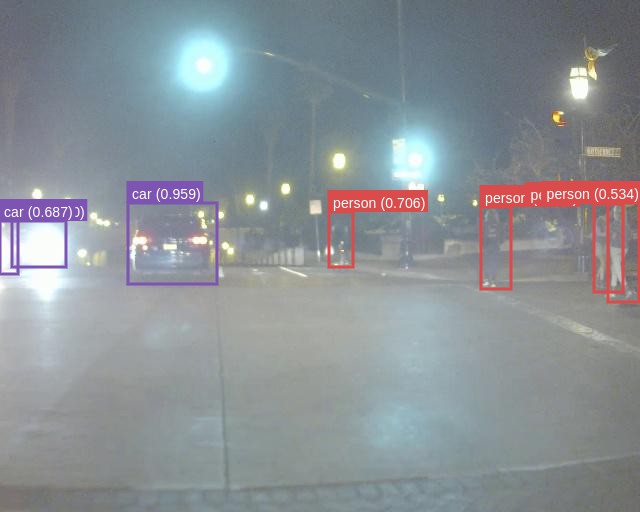}
    \caption{\footnotesize Agnostic CBAM}
    \end{subfigure}
    \begin{subfigure}{0.3\columnwidth}
    \includegraphics[width=1\columnwidth]{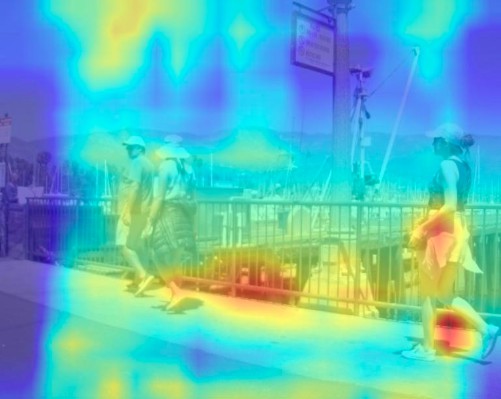}
    \includegraphics[width=1\columnwidth]{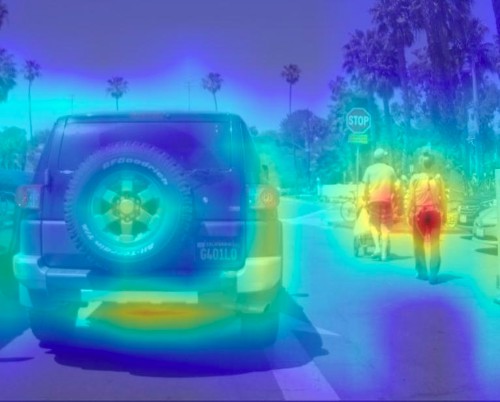}
    \includegraphics[width=1\columnwidth]{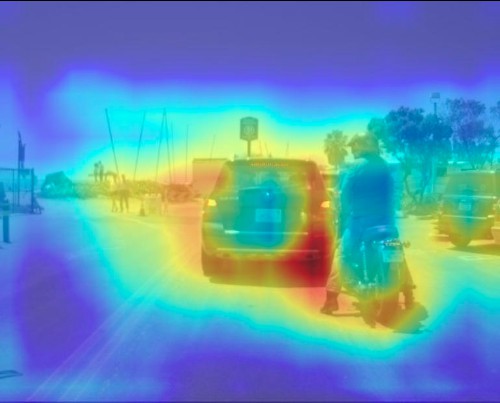}
    \includegraphics[width=1\columnwidth]{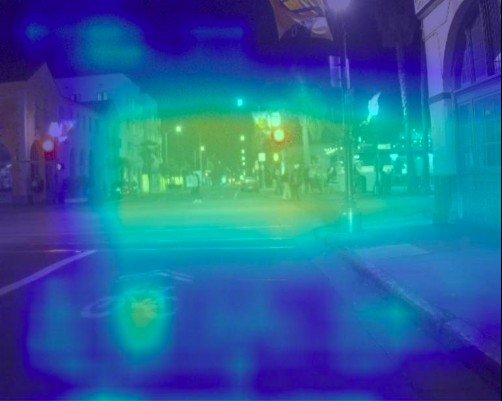}
    \includegraphics[width=1\columnwidth]{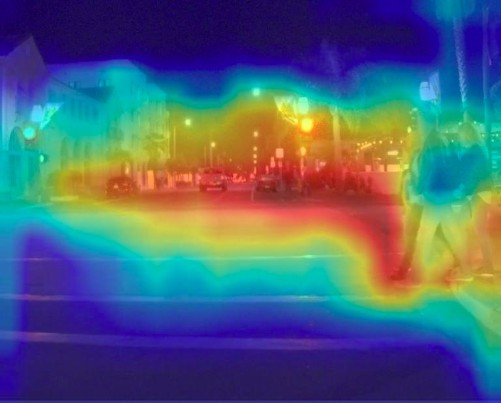}
    \includegraphics[width=1\columnwidth]{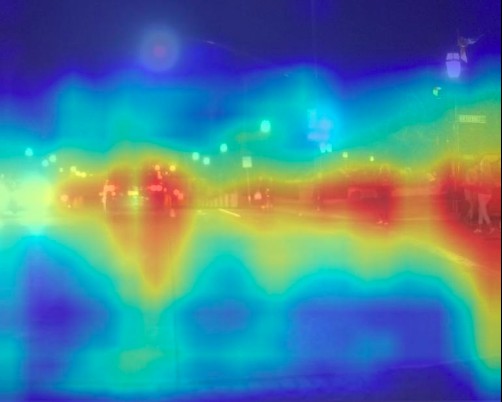}
    \caption{\footnotesize CAM Visual.}
    \end{subfigure}
    \begin{subfigure}{0.3\columnwidth}
    \includegraphics[width=1\columnwidth]{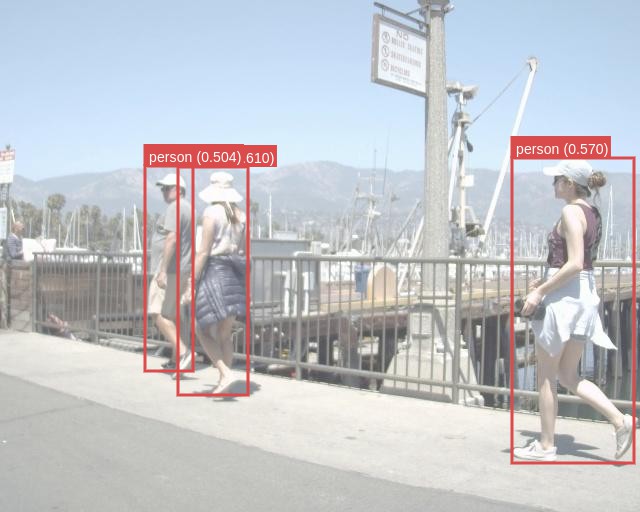}
    \includegraphics[width=1\columnwidth]{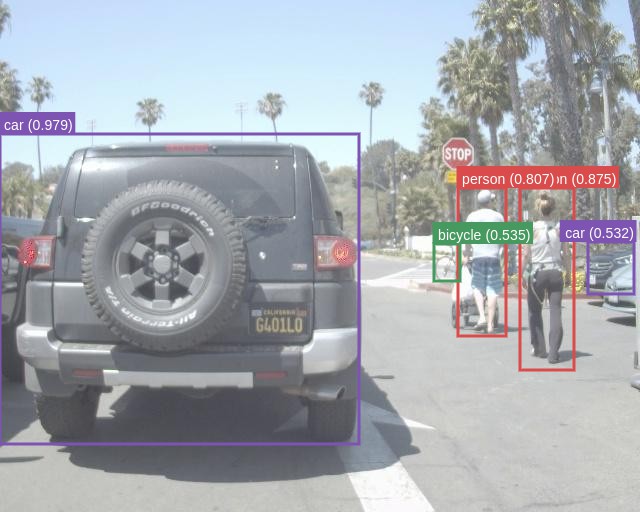}
    \includegraphics[width=1\columnwidth]{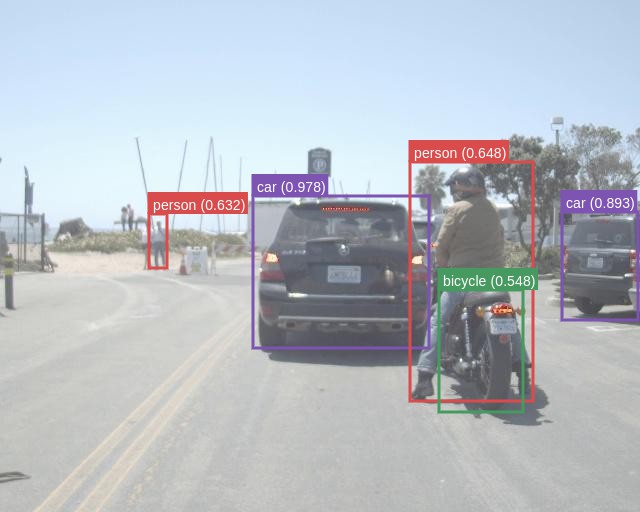}
    \includegraphics[width=1\columnwidth]{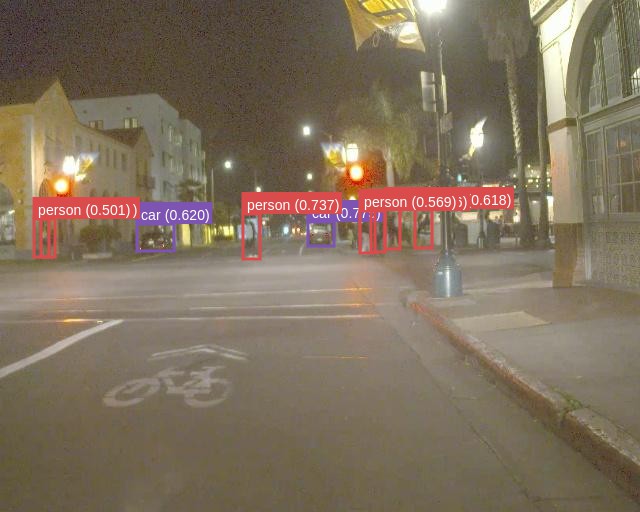}
    \includegraphics[width=1\columnwidth]{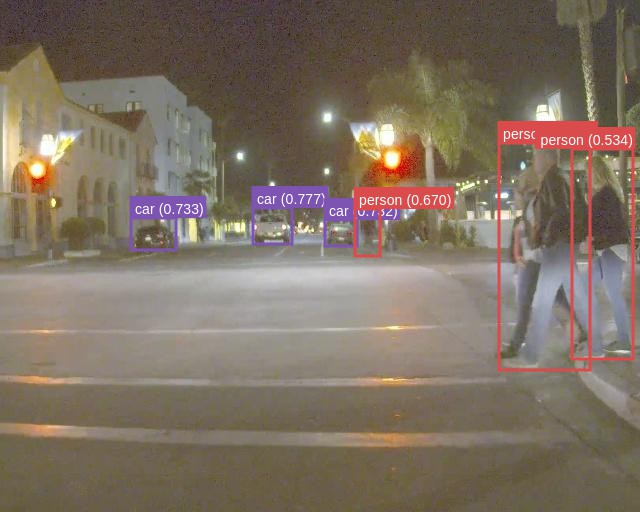}
    \includegraphics[width=1\columnwidth]{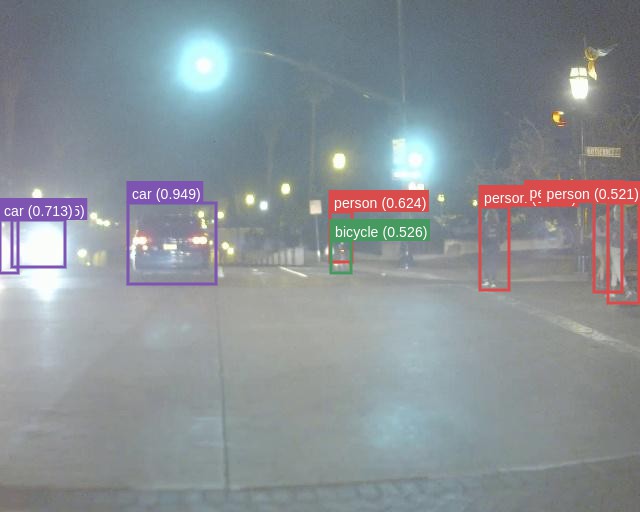}
    \caption{\footnotesize Adaptive CBAM}
    \end{subfigure}
    \begin{subfigure}{0.3\columnwidth}
    \includegraphics[width=1\columnwidth]{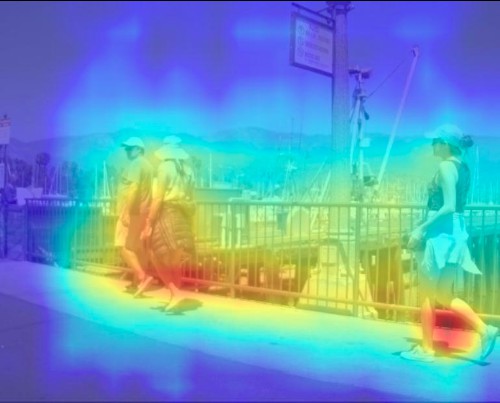}
    \includegraphics[width=1\columnwidth]{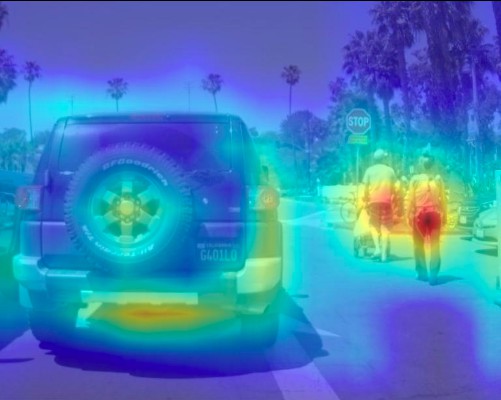}
    \includegraphics[width=1\columnwidth]{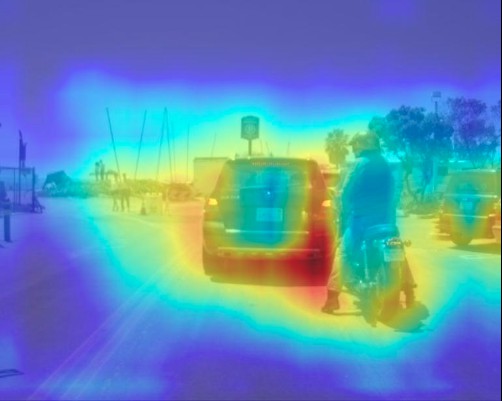}
    \includegraphics[width=1\columnwidth]{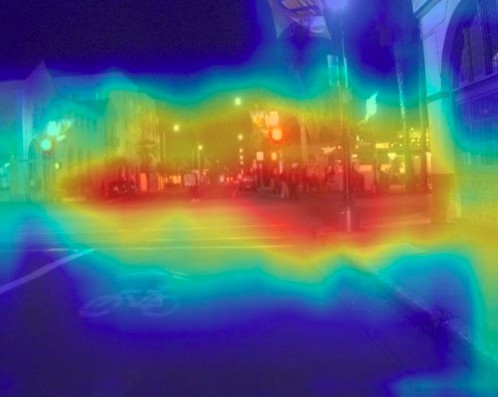}
    \includegraphics[width=1\columnwidth]{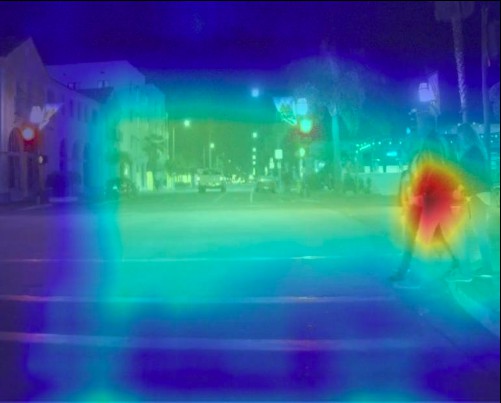}
    \includegraphics[width=1\columnwidth]{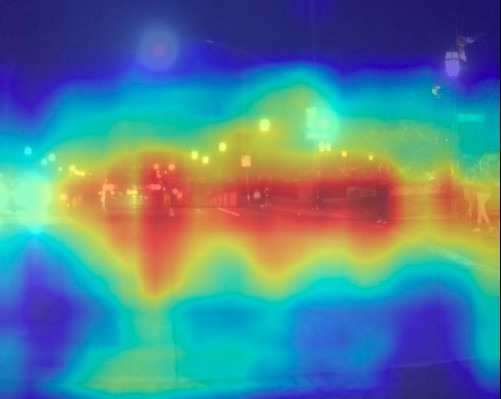}
    \caption{\footnotesize CAM Visual.}
    \end{subfigure}
    \caption{Qualitative detection results on FLIR Aligned dataset with \emph{day} examples in the upper rows and \emph{night} examples in the lower rows. The input RGB and thermal images are overlaid with ground truth (GT) bounding boxes. For each fusion model, we plot the detected bounding boxes and Eigen-CAM~\cite{muhammad2020eigen} visualizations of the CBAM fusion module. (d) and (f) are visualizations of (c) and (e), respectively.}
    \label{fig:results-flir}
\end{figure*}

\begin{figure*}
    \centering
    \begin{subfigure}{0.3\columnwidth}
    \includegraphics[width=1\columnwidth]{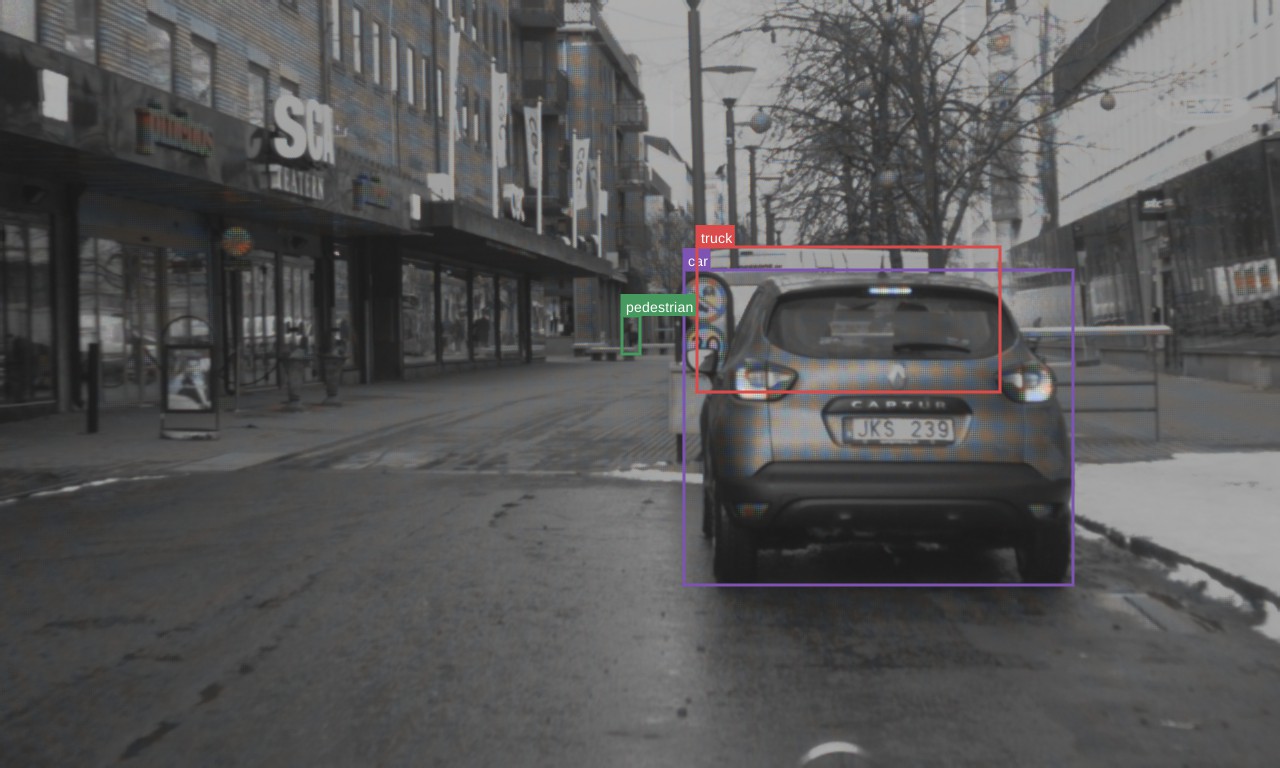}
    \includegraphics[width=1\columnwidth]{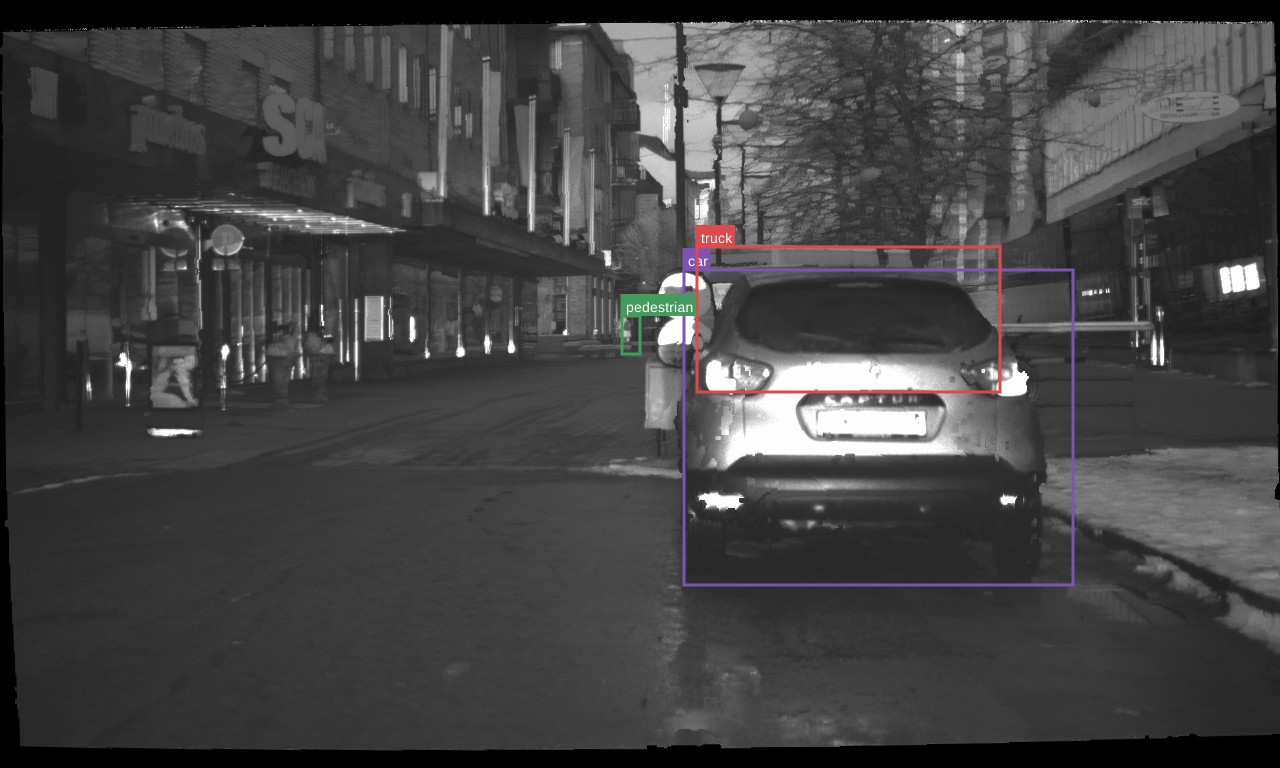}
    \includegraphics[width=1\columnwidth]{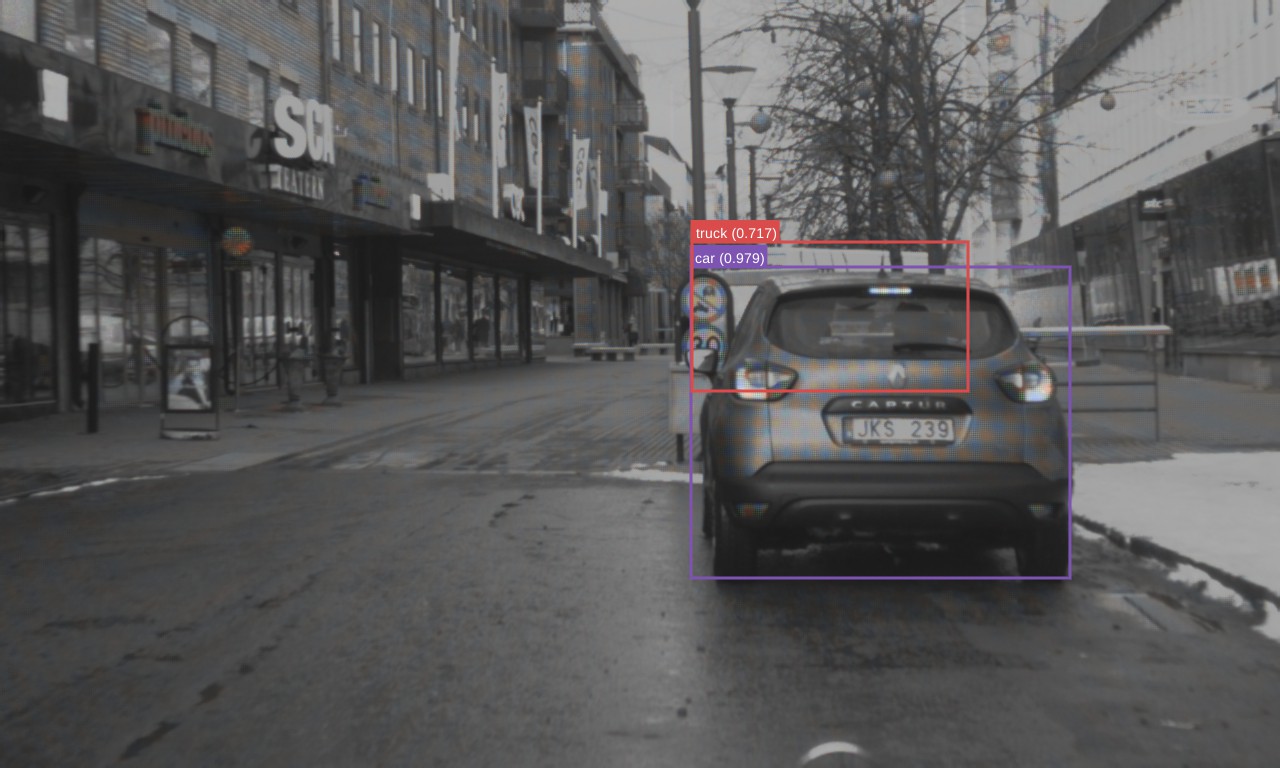}
    \includegraphics[width=1\columnwidth]{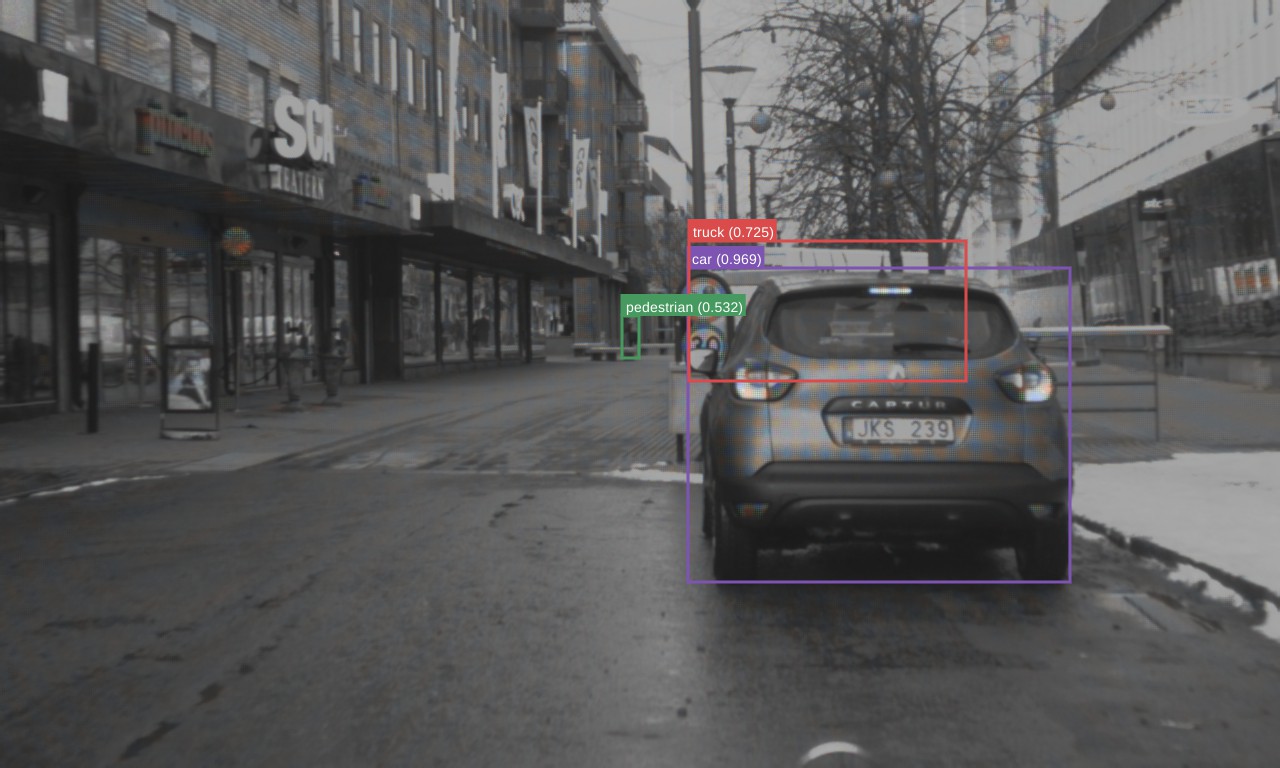}
    \caption{\footnotesize Clear-Day}
    \end{subfigure}
    \begin{subfigure}{0.3\columnwidth}
    \includegraphics[width=1\columnwidth]{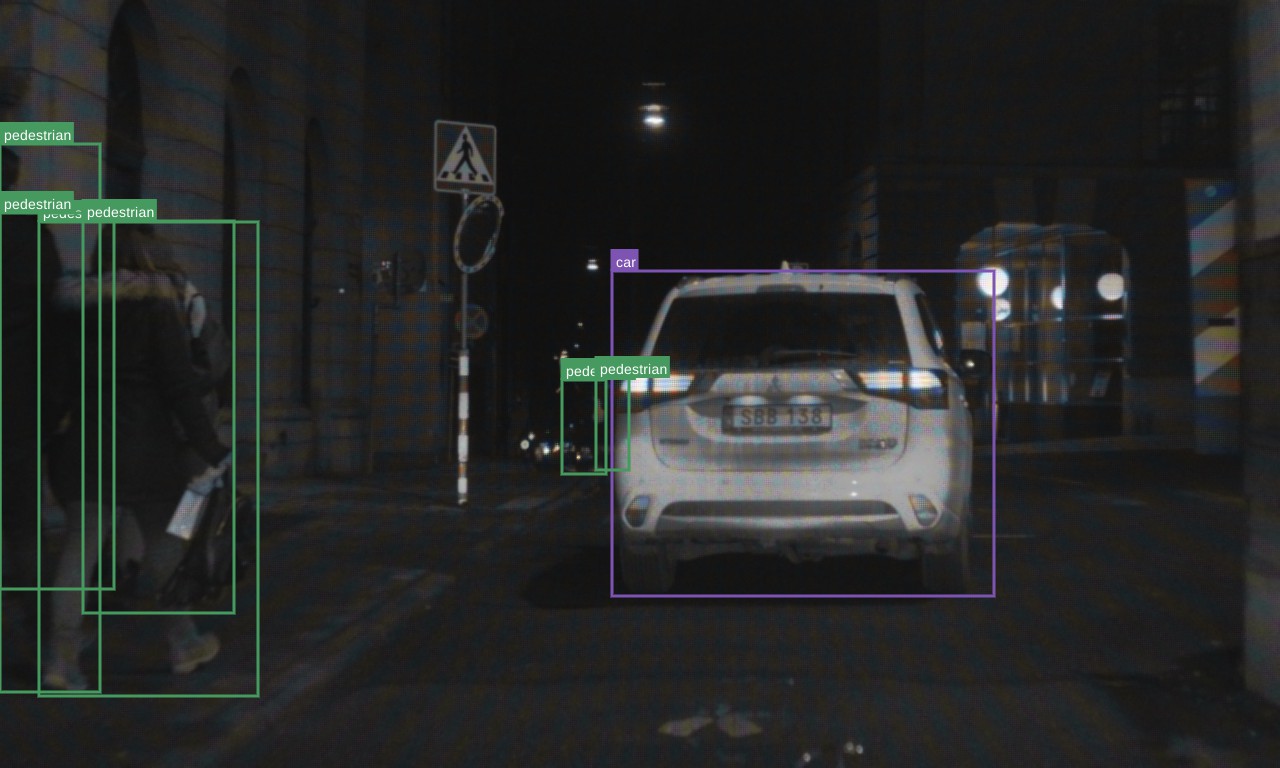}
    \includegraphics[width=1\columnwidth]{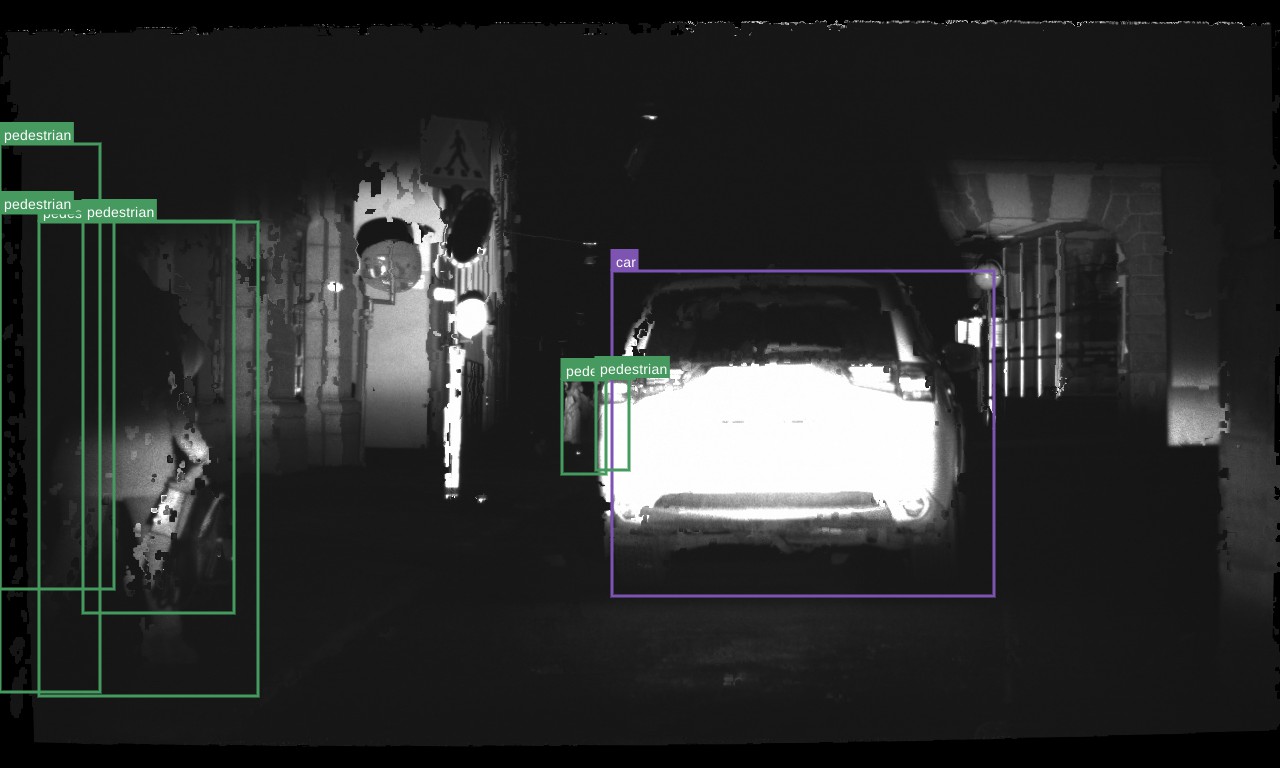}
    \includegraphics[width=1\columnwidth]{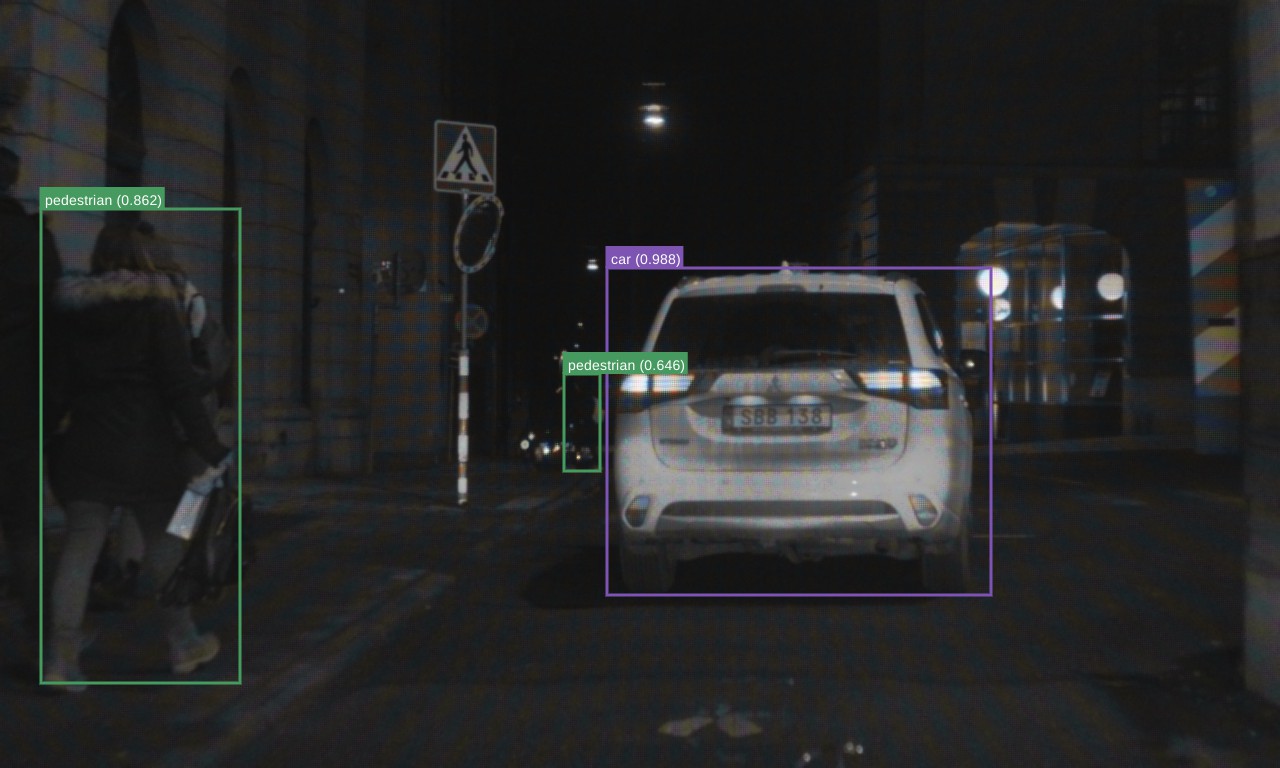}
    \includegraphics[width=1\columnwidth]{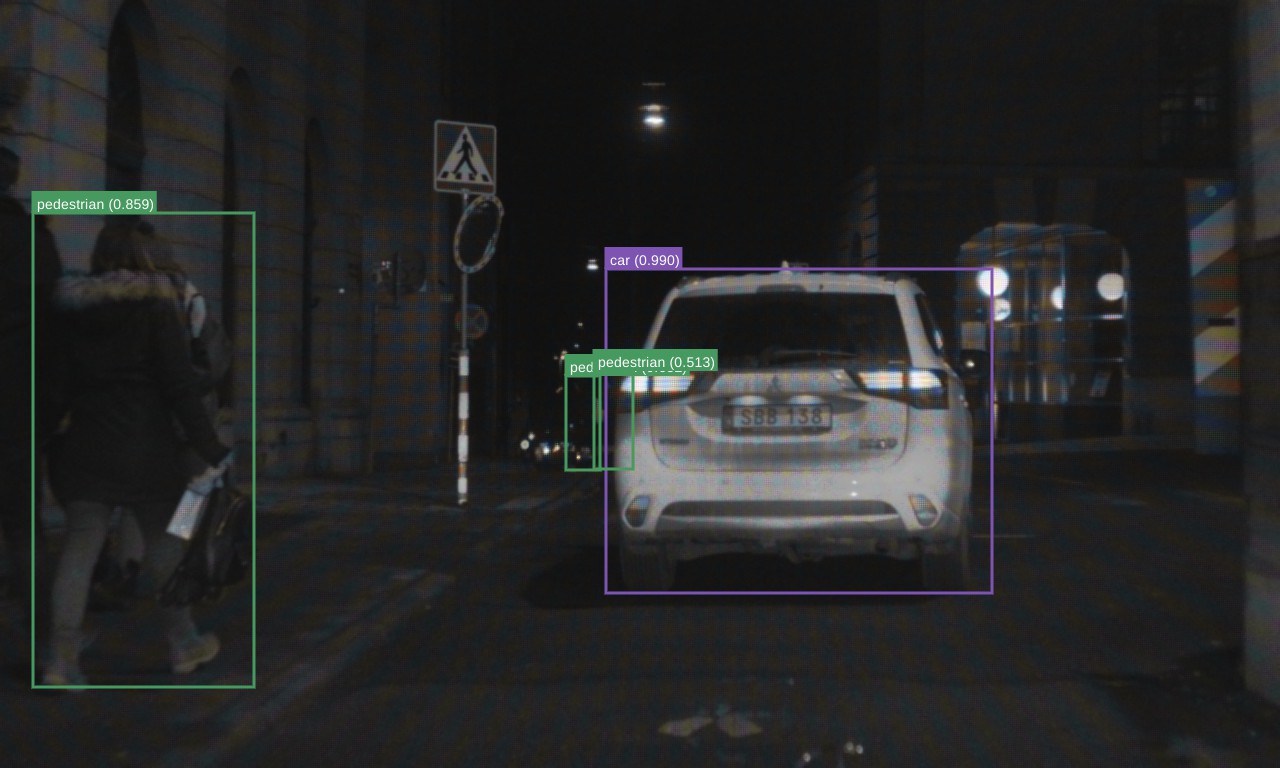}
    \caption{\footnotesize Clear-Night}
    \end{subfigure}
    \begin{subfigure}{0.3\columnwidth}
    \includegraphics[width=1\columnwidth]{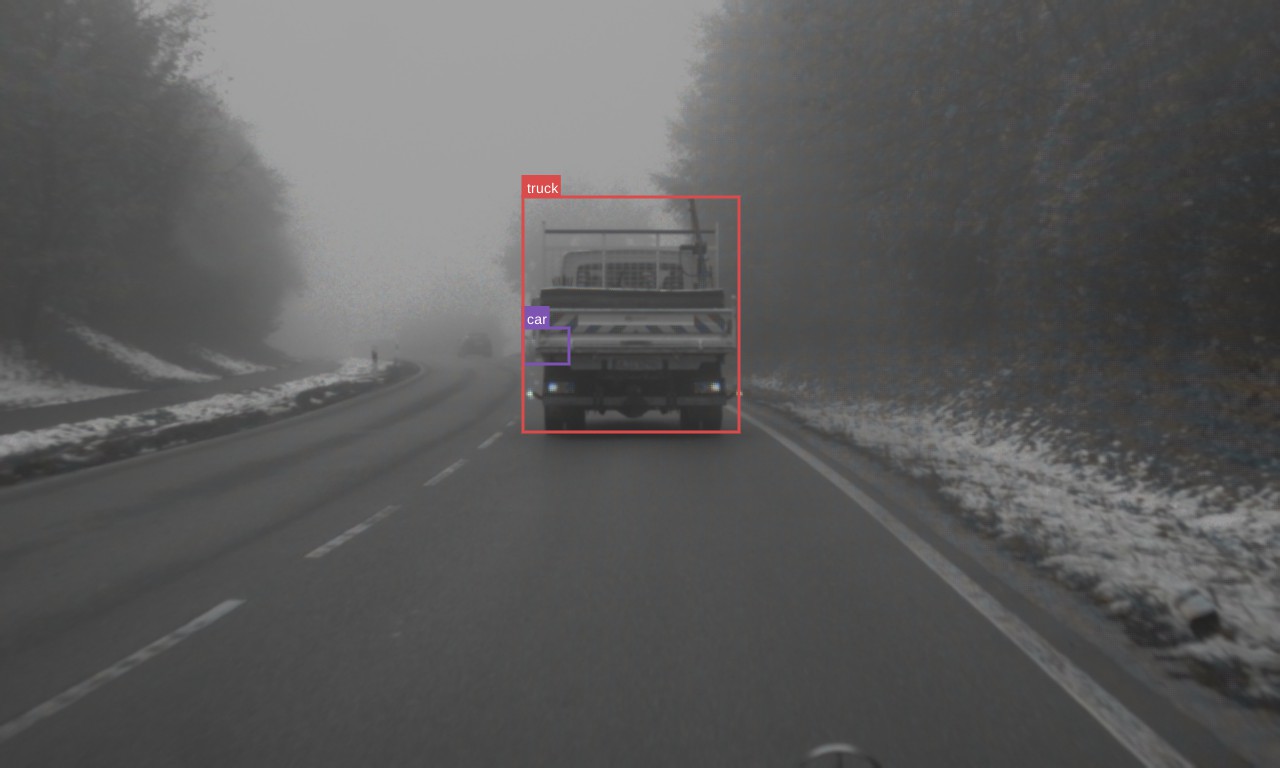}
    \includegraphics[width=1\columnwidth]{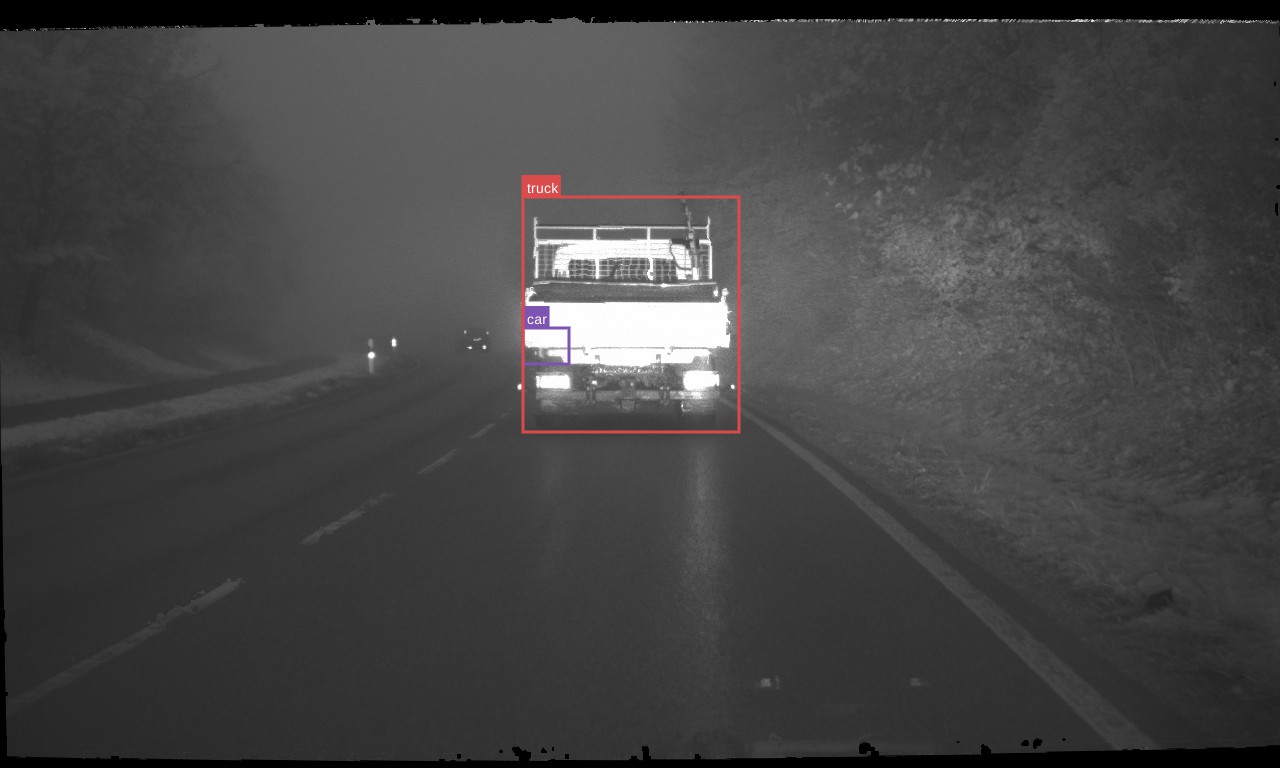}
    \includegraphics[width=1\columnwidth]{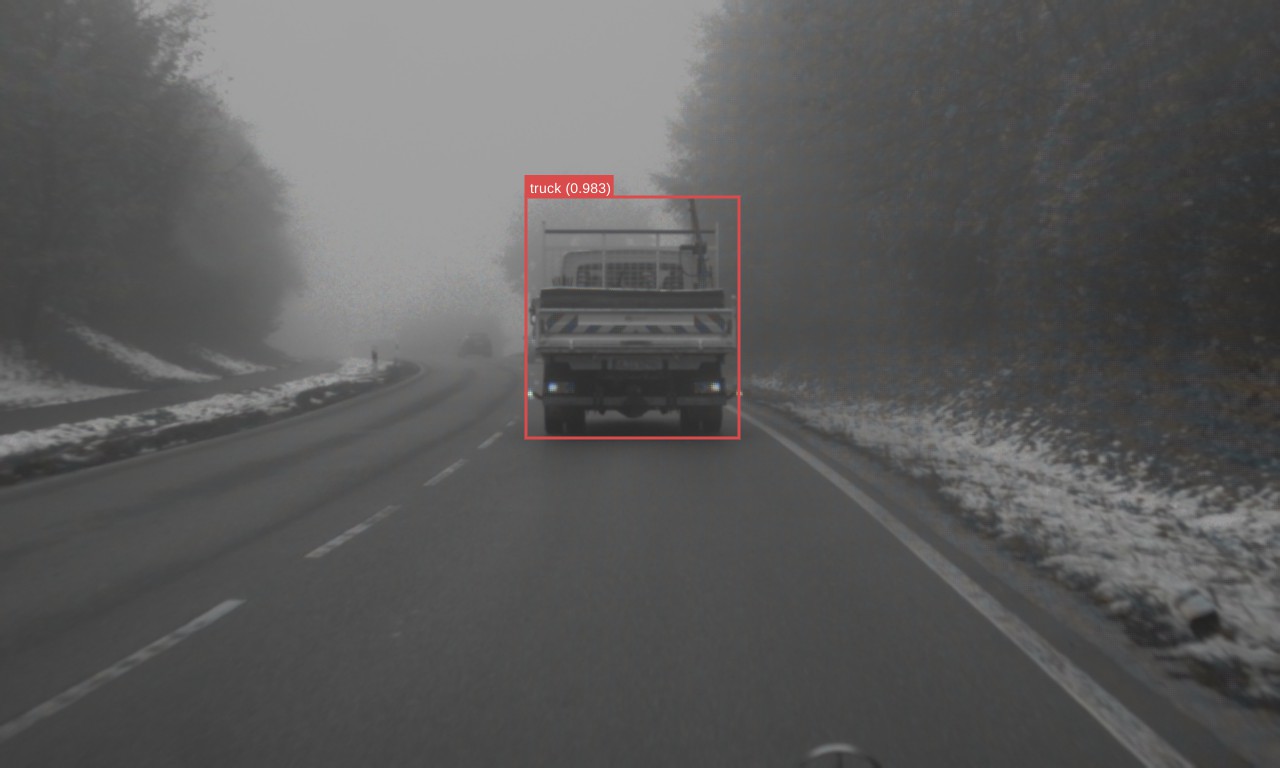}
    \includegraphics[width=1\columnwidth]{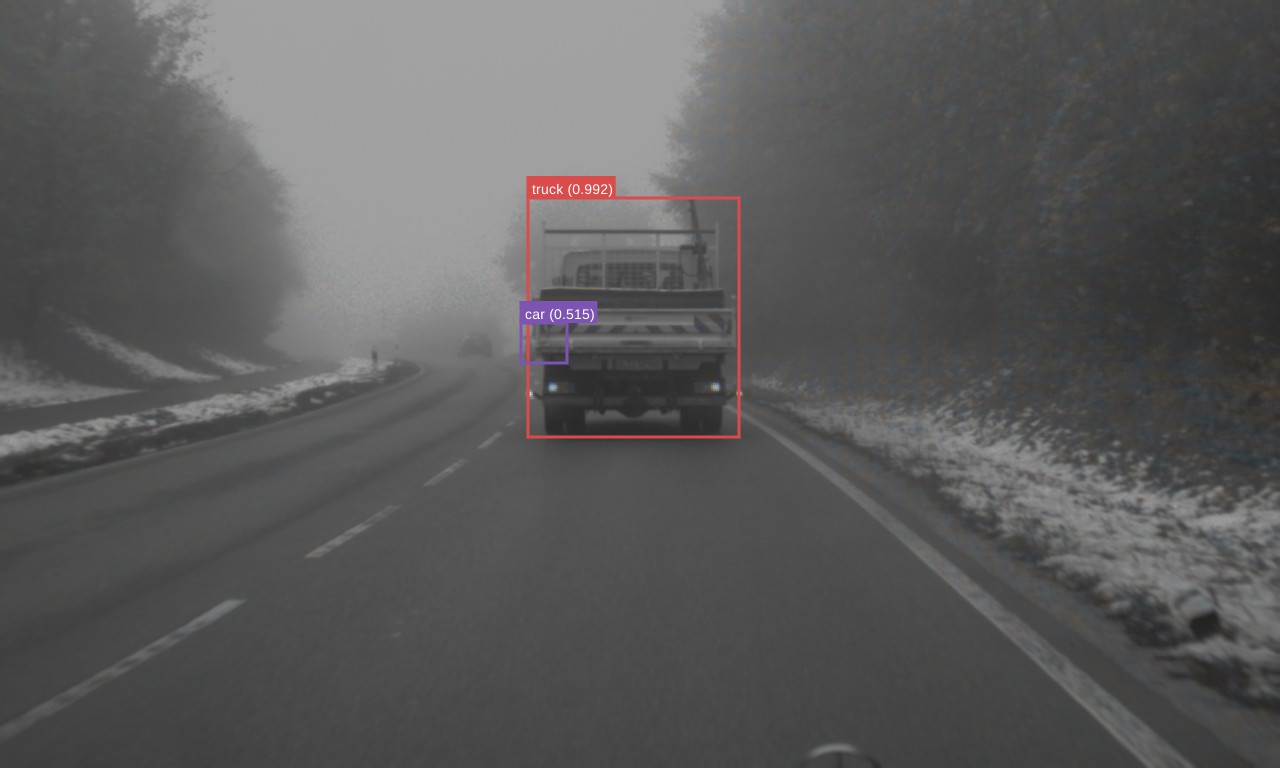}
    \caption{\footnotesize Fog-Day}
    \end{subfigure}
    \begin{subfigure}{0.3\columnwidth}
    \includegraphics[width=1\columnwidth]{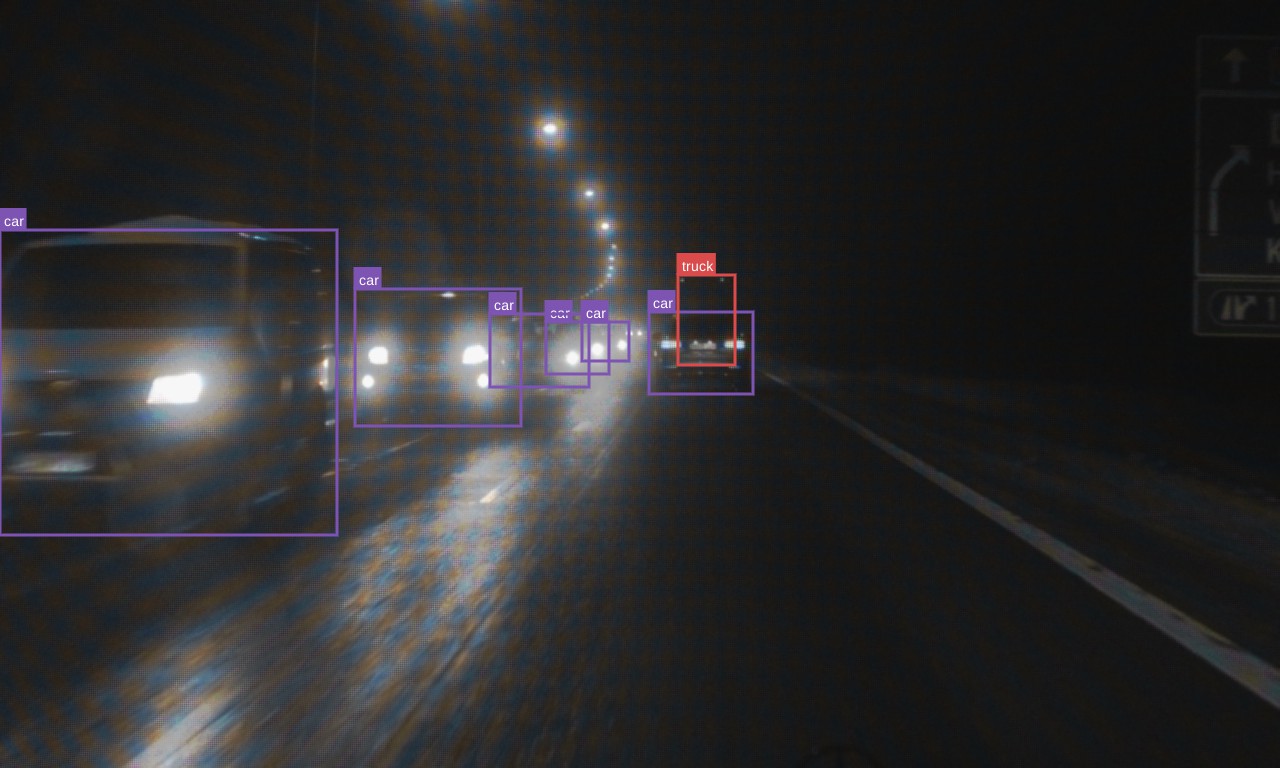}
    \includegraphics[width=1\columnwidth]{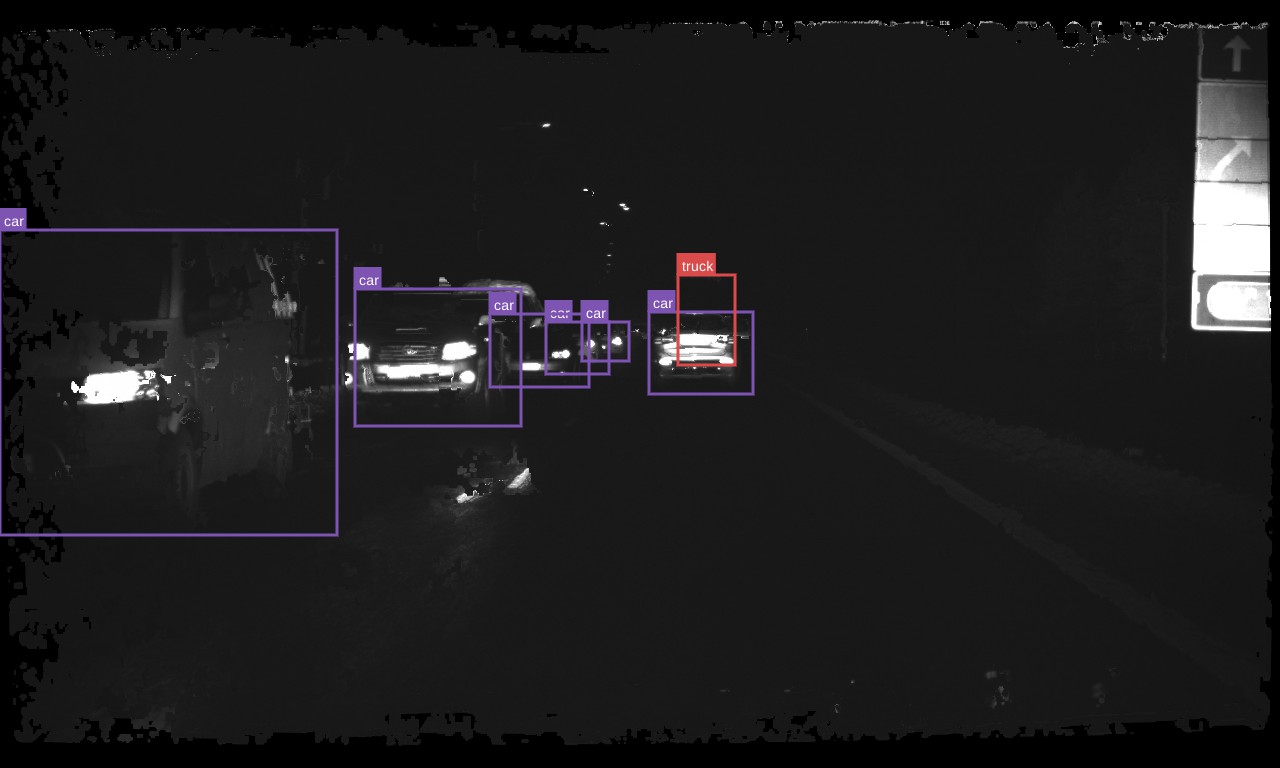}
    \includegraphics[width=1\columnwidth]{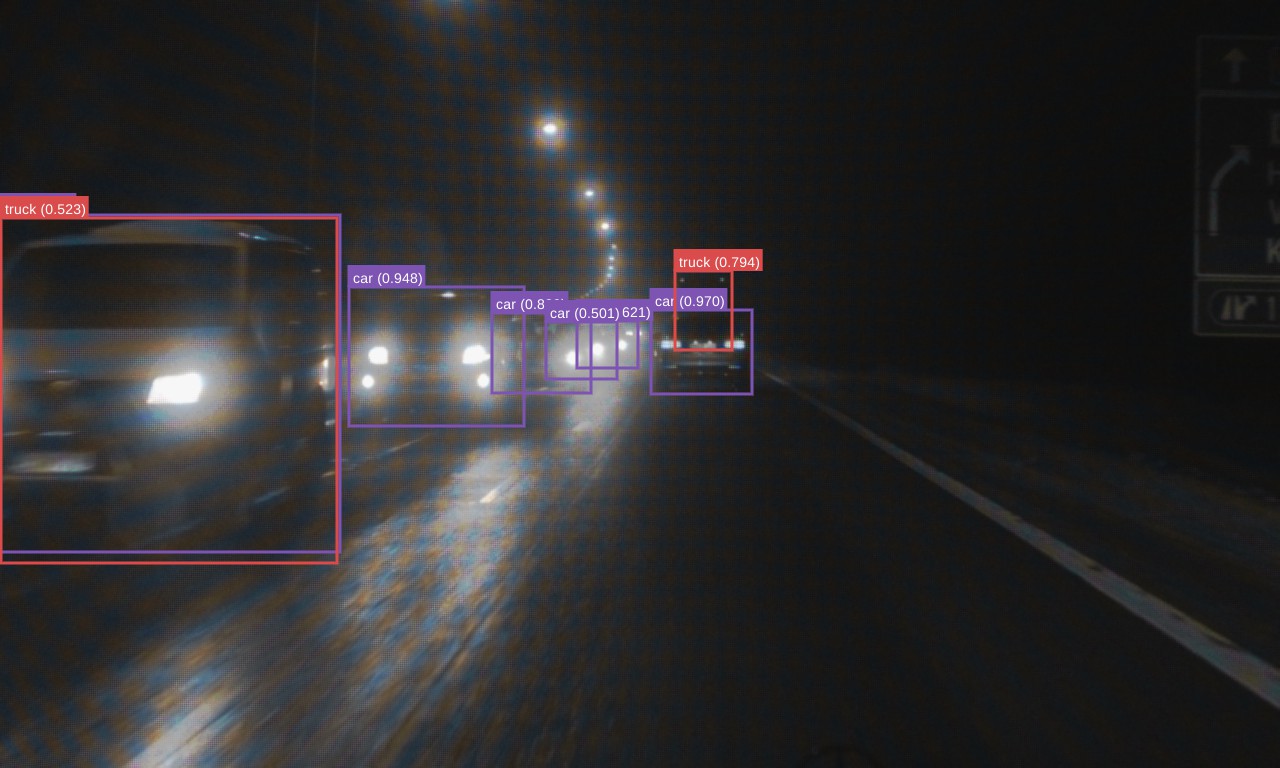}
    \includegraphics[width=1\columnwidth]{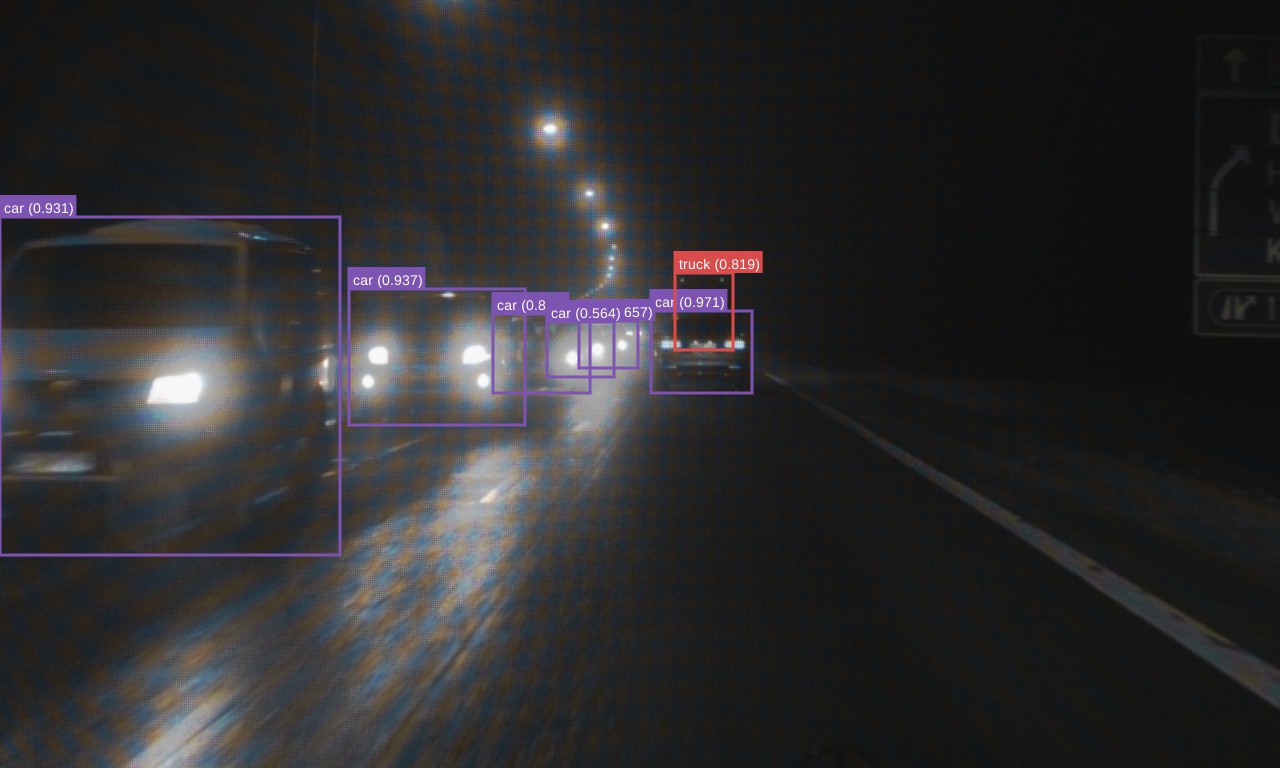}
    \caption{\footnotesize Fog-Night}
    \end{subfigure}
   \begin{subfigure}{0.3\columnwidth}
    \includegraphics[width=1\columnwidth]{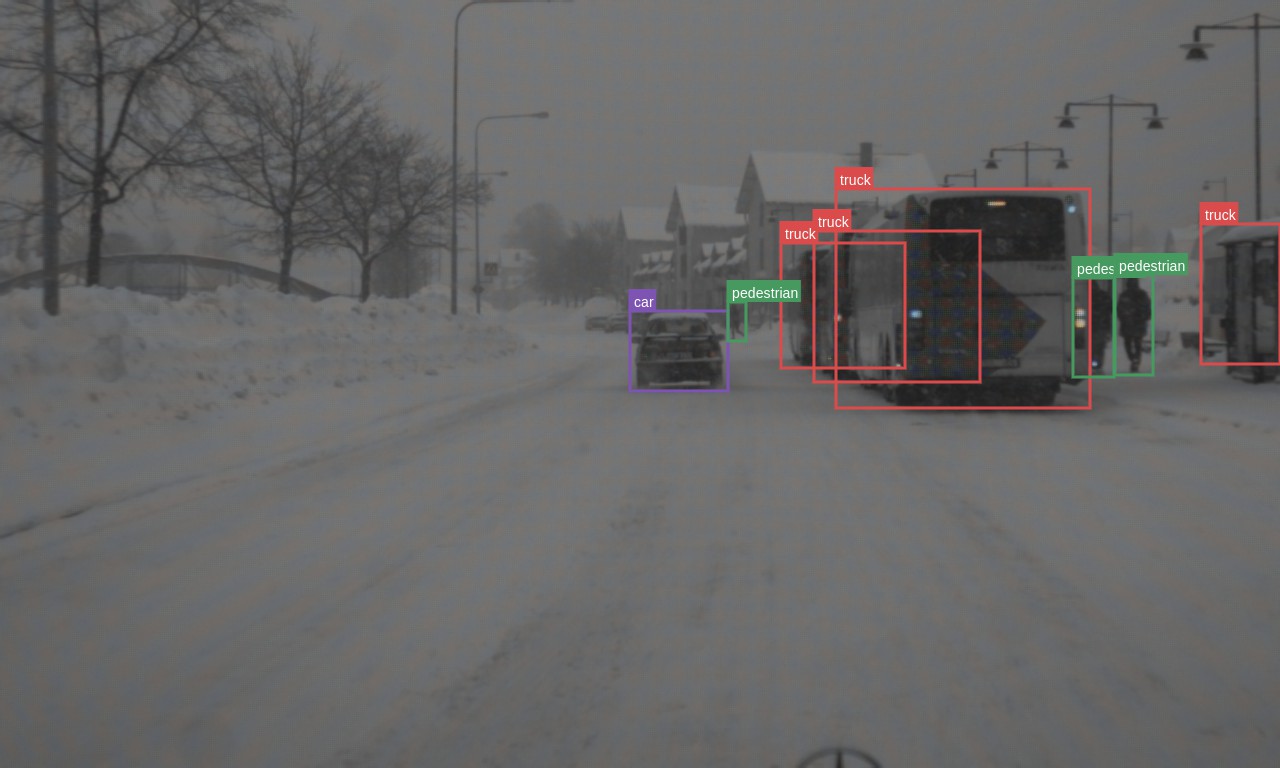}
    \includegraphics[width=1\columnwidth]{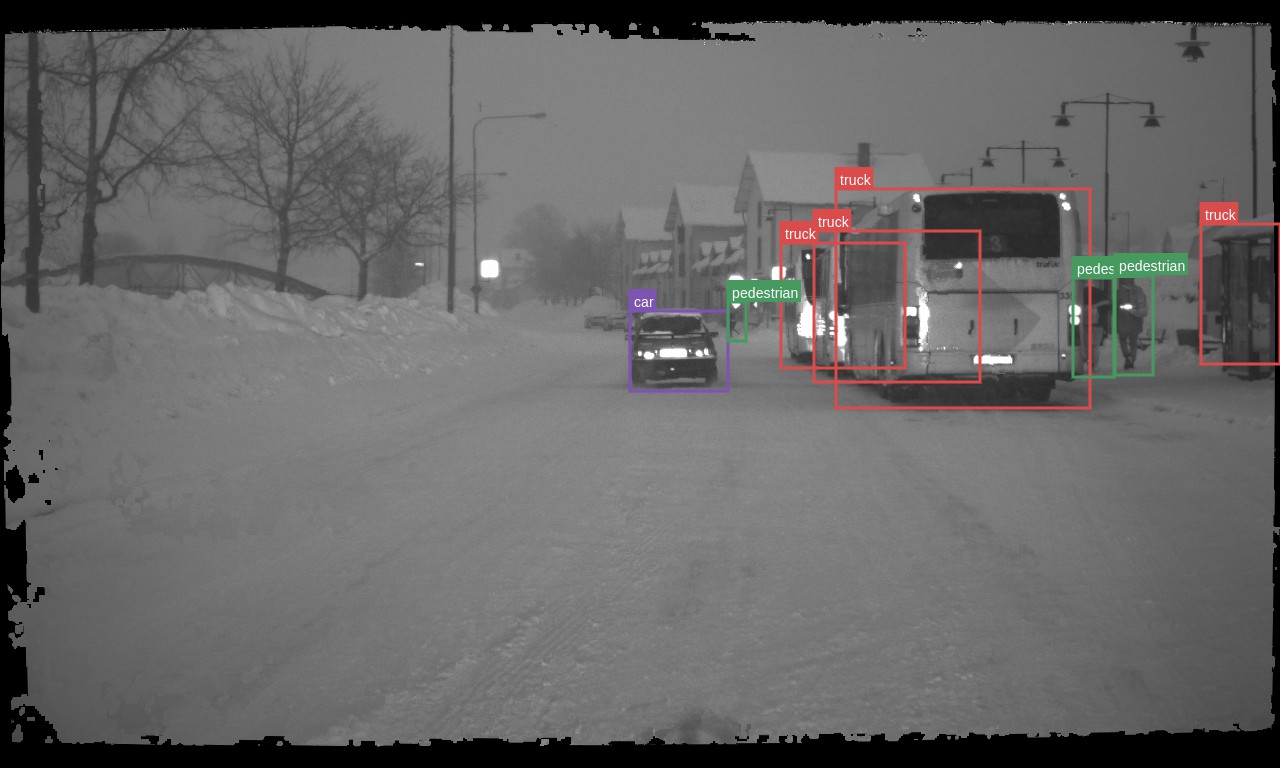}
    \includegraphics[width=1\columnwidth]{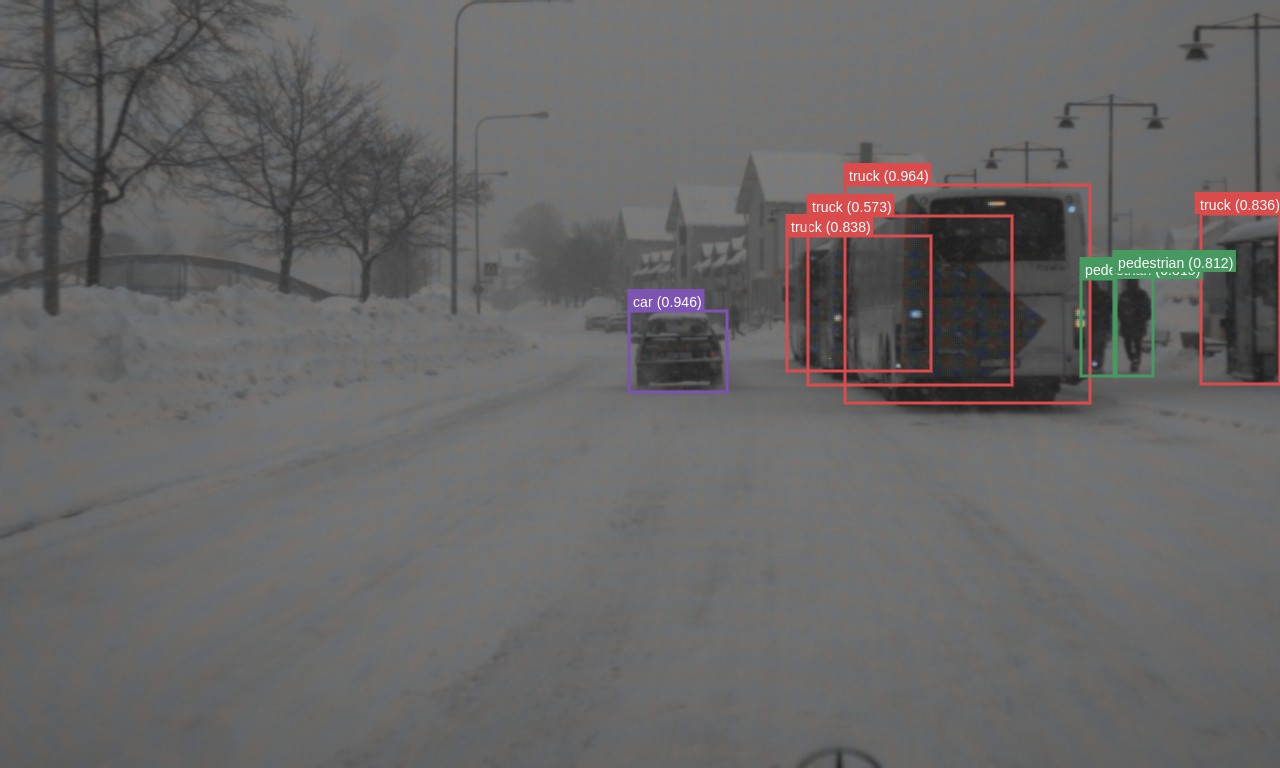}
    \includegraphics[width=1\columnwidth]{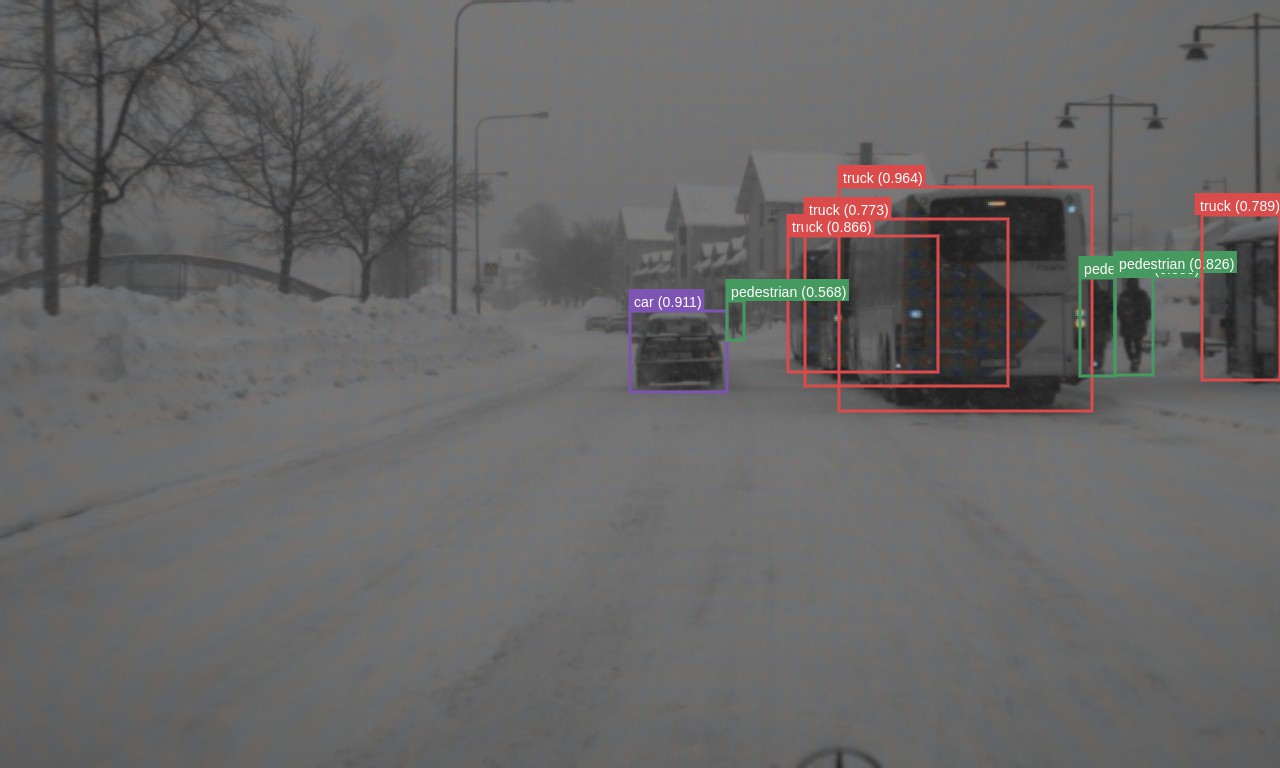}
    \caption{\footnotesize Snow-Day}
    \end{subfigure}
   \begin{subfigure}{0.3\columnwidth}
    \includegraphics[width=1\columnwidth]{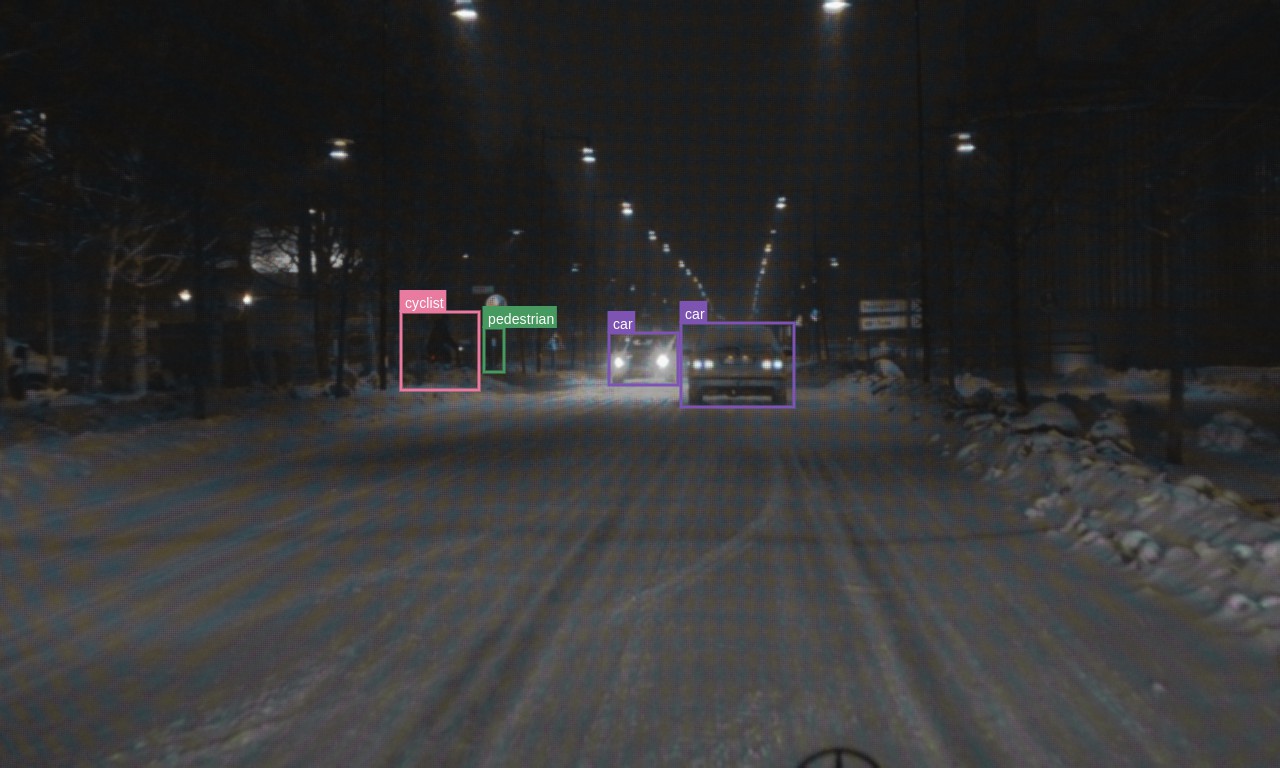}
    \includegraphics[width=1\columnwidth]{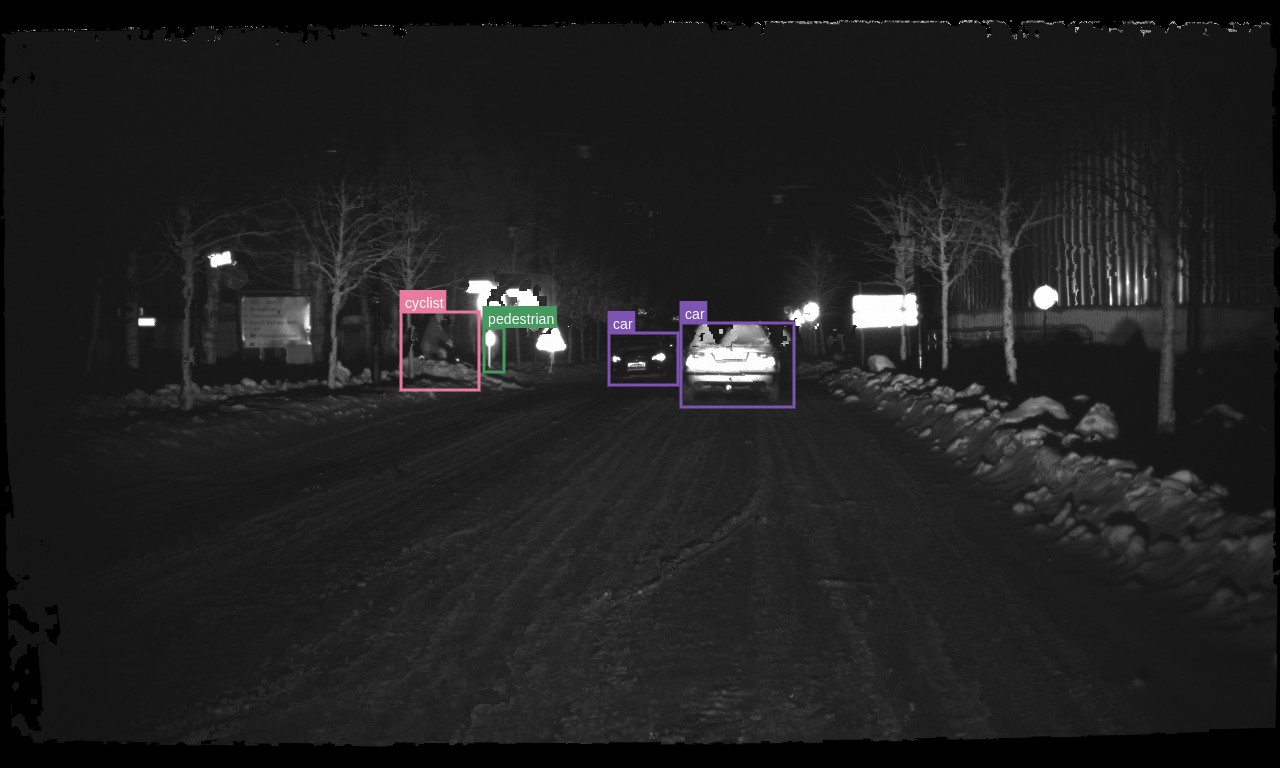}
    \includegraphics[width=1\columnwidth]{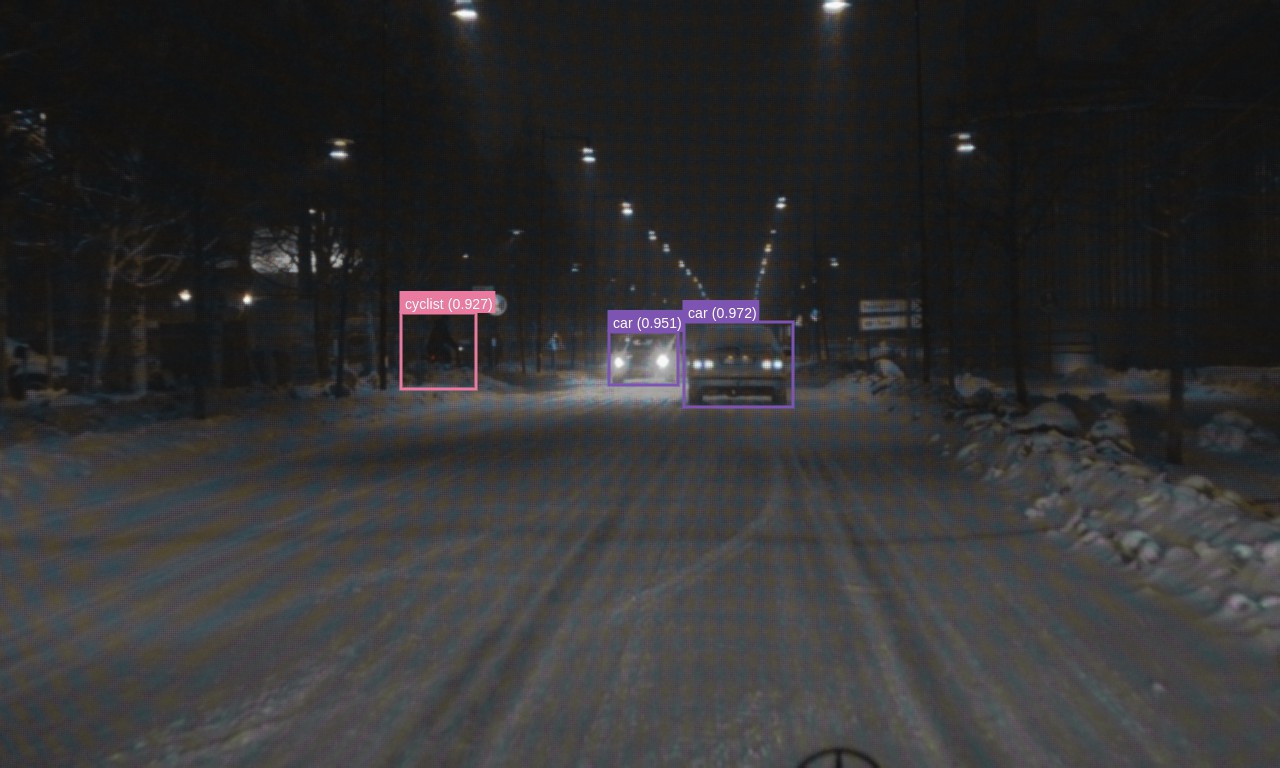}
    \includegraphics[width=1\columnwidth]{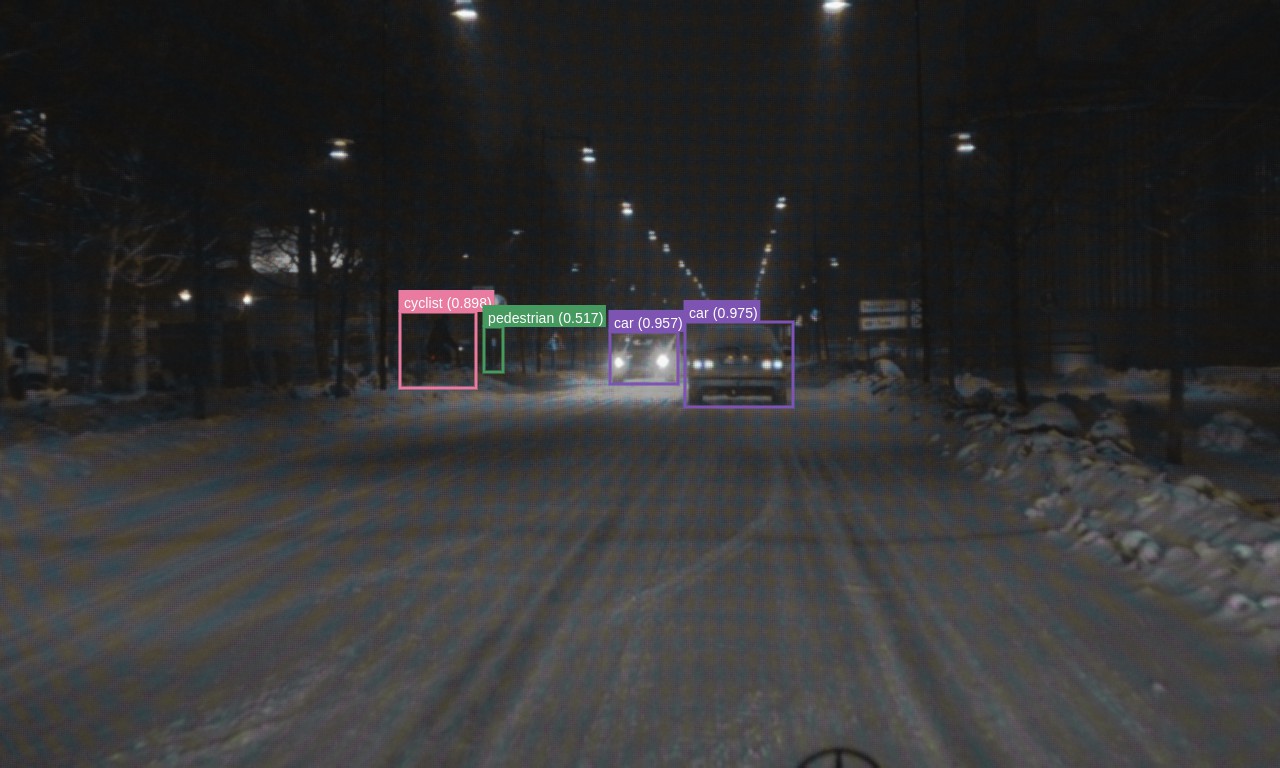}
    \caption{\footnotesize Snow-Night}
    \end{subfigure}
    
    \caption{Object detection results on STF dataset in various scene conditions. From top to bottom: RGB images, gated images, scene-agnostic CBAM detections, and scene-adaptive CBAM detections. RGB and gated images are overlaid with ground truth bounding boxes.}
    \label{fig:results-visual-stf}
\end{figure*}

\subsection{Performance Evaluation}
\label{subsec:performance}
In this section, we validate the proposed method on the three datasets for RGB-X object detection. Auxiliary scene classification is employed to adaptively select suitable fusion modules per input image.

\textbf{Auxiliary Scene Classification:} We train our scene classifiers using ground truth scene labels provided in~\cref{tab:data-splits} by minimizing the standard cross-entropy loss for image classification. Top-1 accuracy of the scene classification is reported in~\cref{tab:cls}, where the classifier attains high accuracy for categorizing various scenes in M$^3$FD, FLIR and STF-Clear (the subset of STF dataset consists of \emph{clear-day} and \emph{clear-night}) datasets. The classifier does not perform as high for STF-Full, possibly because a large portion of \emph{snow} images are also foggy and confused the classifier.

\begin{table}
    \centering
    \caption{Top-1 Accuracy ($\%$) of our scene classifier on the test set of three datasets.}
    \resizebox{0.75\columnwidth}{!}{
    \begin{tabular}{c|c|c|c|c}
    \hlineB{3}
    \multirow{2}{*}{Dataset}  & \multirow{2}{*}{M$^3$FD} & \multirow{2}{*}{FLIR} & \multicolumn{2}{c}{Seeing Through Fog} \\ \cline{4-5}
       & &   & Clear & Full \\ 
    \hline
    Accuracy & 91.42 & 96.35 & 96.01 & 77.02 \\
    \hlineB{3}
    \end{tabular}
    }
    \label{tab:cls}
\end{table}

\begin{table}
\centering
    \caption{Object detection results (mAP@0.5) and speed (s) on M$^3$FD dataset. Due to the difference in scene splits between baselines and our models, only results on the \emph{full} test set are comparable across all methods.}
    \label{tab:result-m3fd}
    \resizebox{\columnwidth}{!}{
        \begin{tabular}{l|cccc|c|c}
            \hlineB{3}
            \multirow{2}{*}{Method} & \multicolumn{5}{c|}{Test Scene} & \multirow{2}{*}{\shortstack[c]{Inference \\ Speed (s)}} \\ \cline{2-6}
            & Day & Night & Overcast & Challenge & Full & \\
            \hlineB{3}
            RGB only  & 71.59 & 91.06 & 81.55 & 80.03 & 77.79 & 0.016 \\
            Thermal only & 65.68 & 89.17 & 79.66 & 76.39 & 74.64 & 0.016 \\
            \hlineB{3}
            U2F~\cite{xu2020u2fusion} & 73.80 & 86.8 & 73.10 & 97.6 & 77.5 & 0.129$^{\dagger}$ \\
            TarDAL~\cite{liu2022target} & 74.50 & 89.30 & 74.10 & 98.30 & 77.80 & 0.047$^{\dagger}$ \\
            EAEFNet~\cite{liang2023explicit} & 78.30 & 89.50 & 78.60  & 97.90 & 80.10 & ---  \\
            \cline{2-5}
            Scene-Agnostic CBAM \textbf{(ours)} & 74.53 & \bf 93.09 & 84.11 & 81.06 & 80.46 & 0.028       \\
            Scene-Adaptive CBAM \textbf{(ours)} & \bf 75.92 & 92.55 & \bf 85.14 & \bf 82.72 & \bf 81.46 & 0.032        \\
            \hlineB{3}
        \end{tabular}
    }
    \captionsetup{font=footnotesize}
    \caption*{$^{\dagger}$ Includes image fusion and object detection inference time.}
\end{table}

\begin{table}
    \centering
        \caption{Object detection results and speed (s) on FLIR Aligned dataset. AP@0.5 for each object category is reported.}
    \label{tab:flir-results}
    \resizebox{\columnwidth}{!}{\begin{tabular}{l|ccc|ccc|c}
        \hlineB{3}
        \multirow{2}{*}{Method} & \multirow{2}{*}{Person} & \multirow{2}{*}{Bicycle} & \multirow{2}{*}{Car} & \multirow{2}{*}{mAP@0.5} & \multirow{2}{*}{mAP@0.75} & \multirow{2}{*}{mAP$^{\dagger}$} & \multirow{2}{*}{\shortstack[c]{Inference \\ Speed (s)}} \\ &&&&&&&\\
        \hlineB{3}
        RGB only                                & 60.79 & 37.25 & 73.94 & 57.32 & 17.6 & 24.7 & 0.016 \\
        Thermal only                            & 82.86 & 50.80 & 82.83 & 72.16 & 33.4 & 37.0 & 0.016 \\
        \hlineB{3}
        GAFF~\cite{zhang2021guided}             & 76.60 & 59.40 & 85.50 & 72.9 & 32.9 & 37.5 & 0.061\\
        CFR\_3\cite{zhang2020multispectral}     & 74.49 & 57.77 & 84.91 & 72.93 & --- & --- & 0.050 \\
        RetinaNet + MFPT\cite{zhu2023multi}     & 78.1 & 65.0 & 87.3 & 76.80 & --- & --- & 0.050 \\
        UA-CMDet\cite{sun2022drone}             & 83.20 & 64.30 & 88.40 & 78.60 & --- & --- & --- \\
        CFT~\cite{qingyun2021cross}             & --- & --- & --- & 78.7 & 35.5 & 40.2 & 0.026 \\
        CSAA\cite{cao2023multimodal}            & --- & --- & --- & 79.20 & 37.4 & 41.3 & 0.031 \\
        FasterRCNN + MFPT\cite{zhu2023multi}    & 83.2 & 67.7 & 89.0 & 80.00 & --- & --- & 0.080 \\
        LRAF-Net\cite{fu2023lraf}               & --- & --- & --- & 80.50 & --- & 42.8 & --- \\
        Scene-agnostic CBAM \textbf{(ours)}       & 88.26 & 77.43 & 90.68 & 85.45 & \textbf{43.3} & 46.8 & 0.028 \\
        Scene-adaptive CBAM \textbf{(ours)}       & \textbf{88.92} & \textbf{78.61} & \textbf{90.94} & \textbf{86.16} & 43.0 & \textbf{47.1} & 0.032 \\
        \hlineB{3}
    \end{tabular}}
    \captionsetup{font=footnotesize}
    \caption*{$^{\dagger}$ mAP refers to mAP@0.5:0.95}
\end{table}

\begin{figure}
    \centering
    \begin{subfigure}{0.245\columnwidth}
    \includegraphics[width=1\columnwidth]{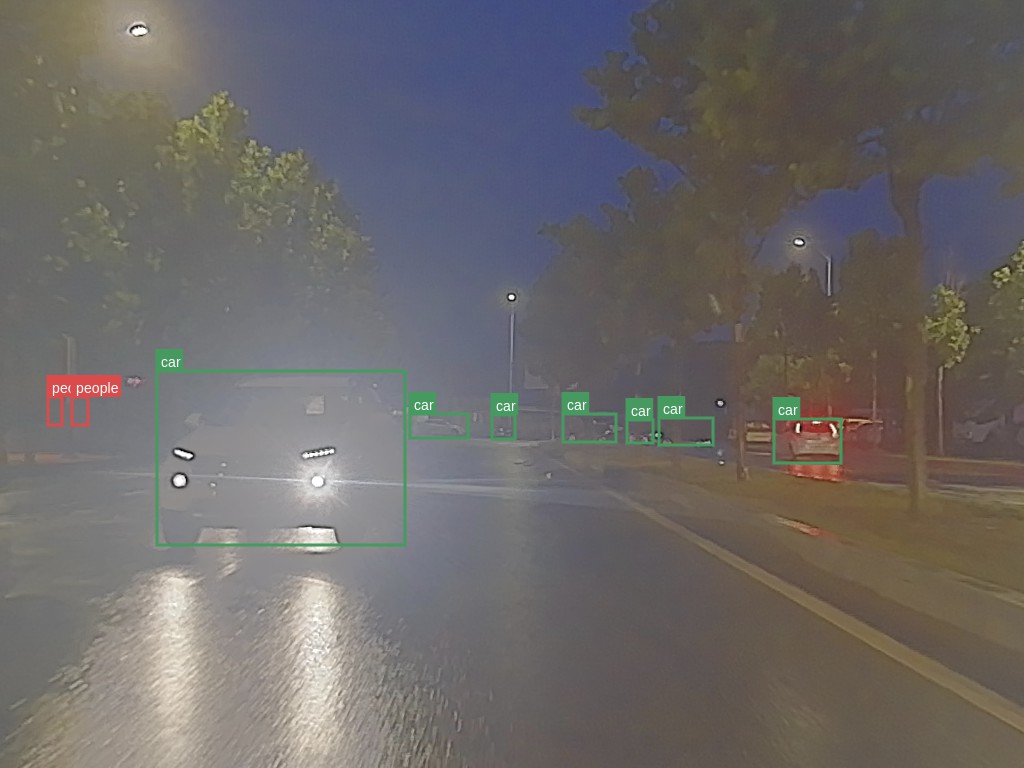} 
    \includegraphics[width=1\columnwidth]{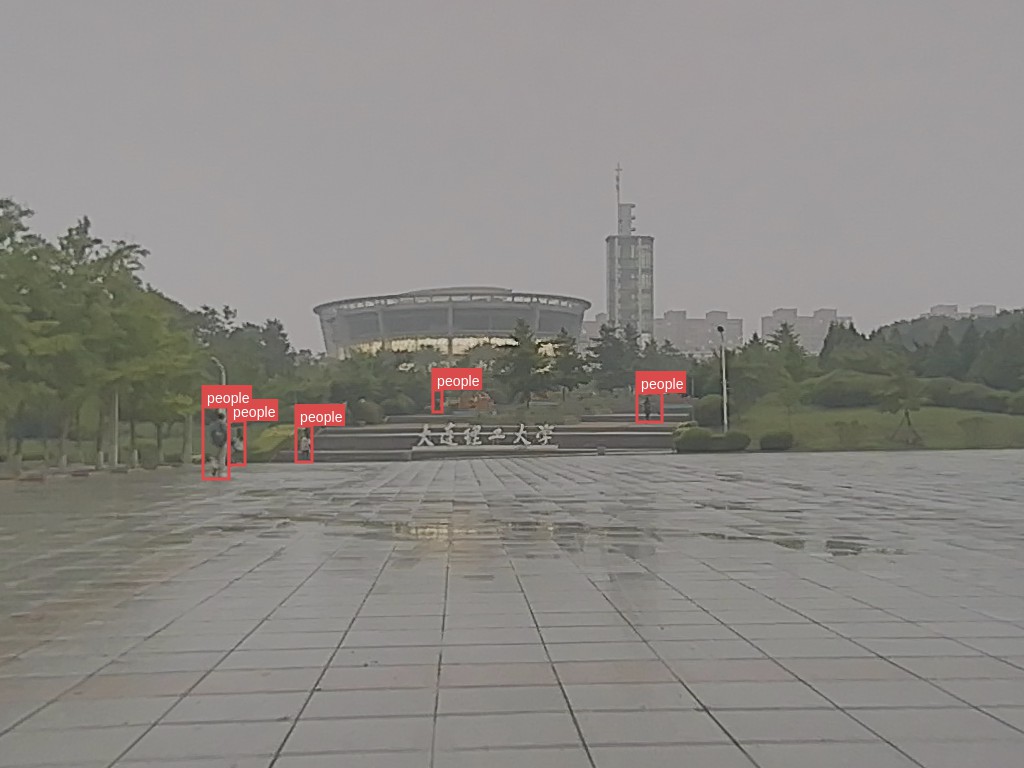} 
    \includegraphics[width=1\columnwidth]{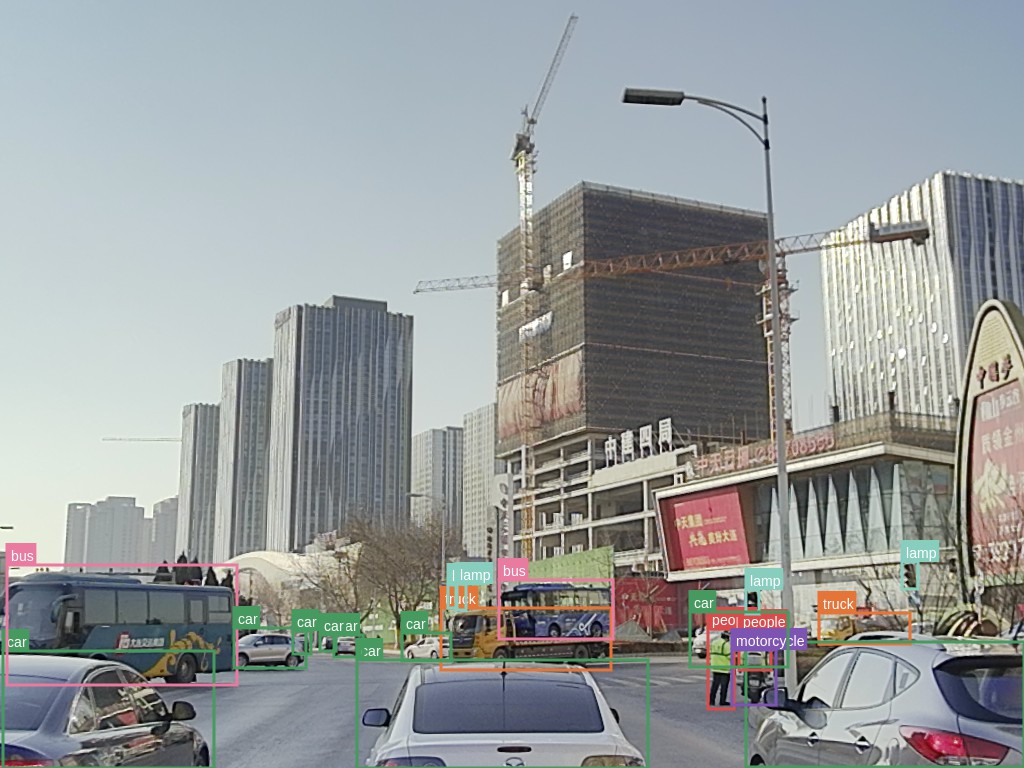}
    \caption{\scriptsize RGB-GT}
    \end{subfigure}
    \hspace{-1.5mm}
    \begin{subfigure}{0.245\columnwidth}
    \includegraphics[width=1\columnwidth]{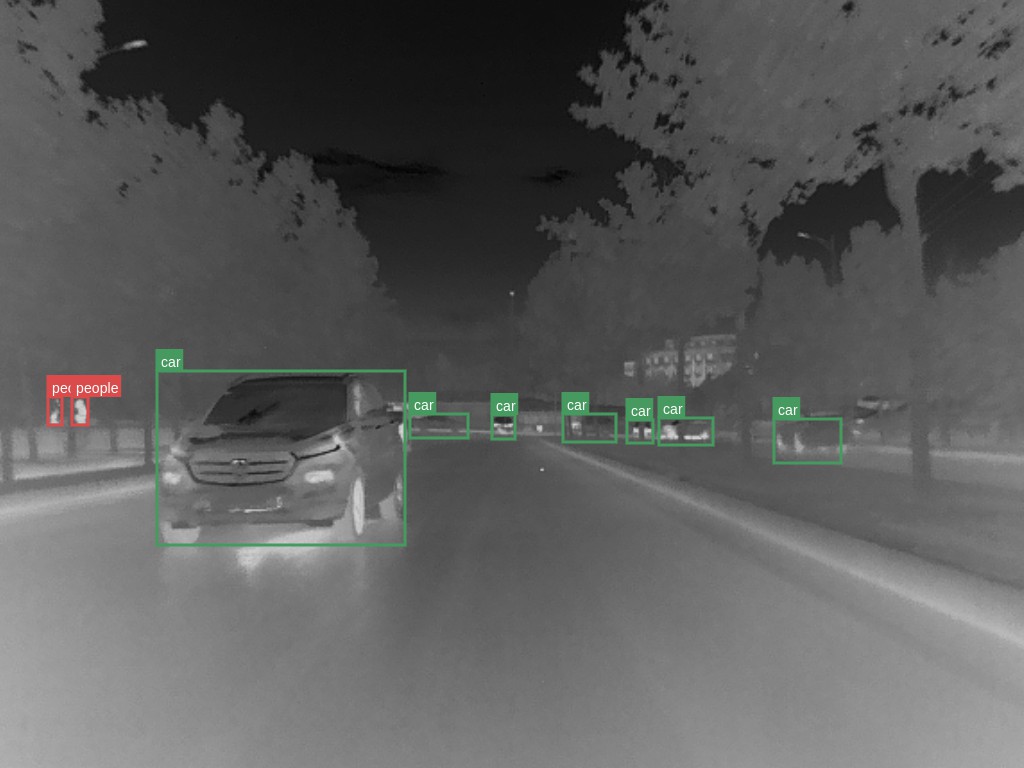} 
    \includegraphics[width=1\columnwidth]{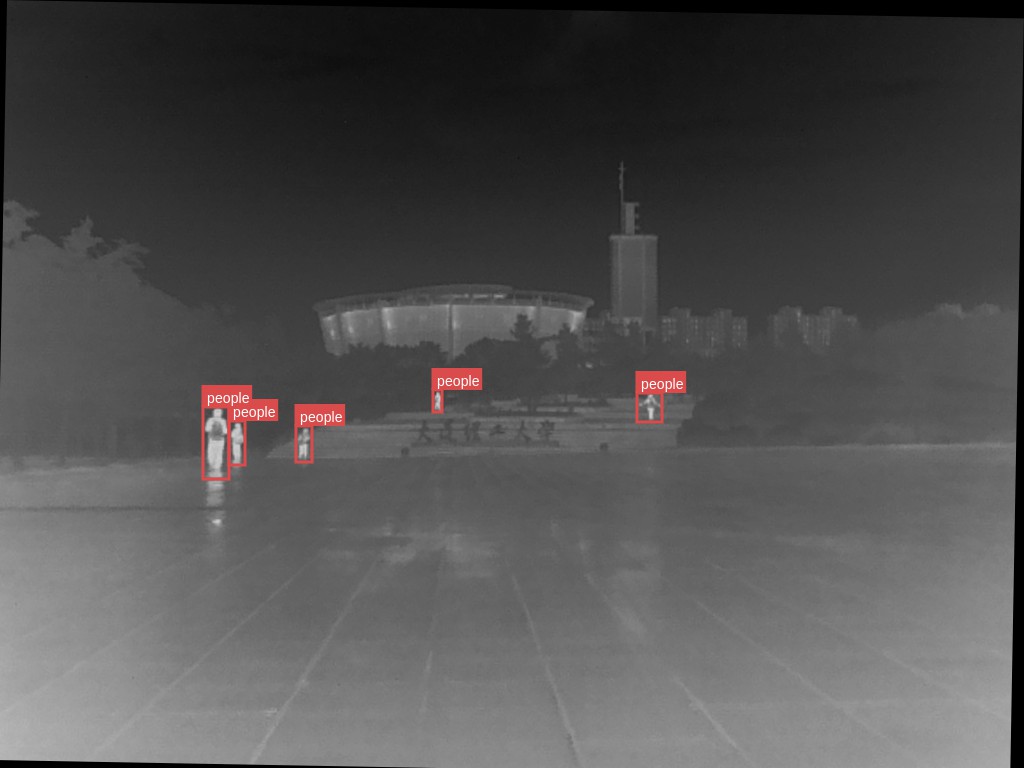}
    \includegraphics[width=1\columnwidth]{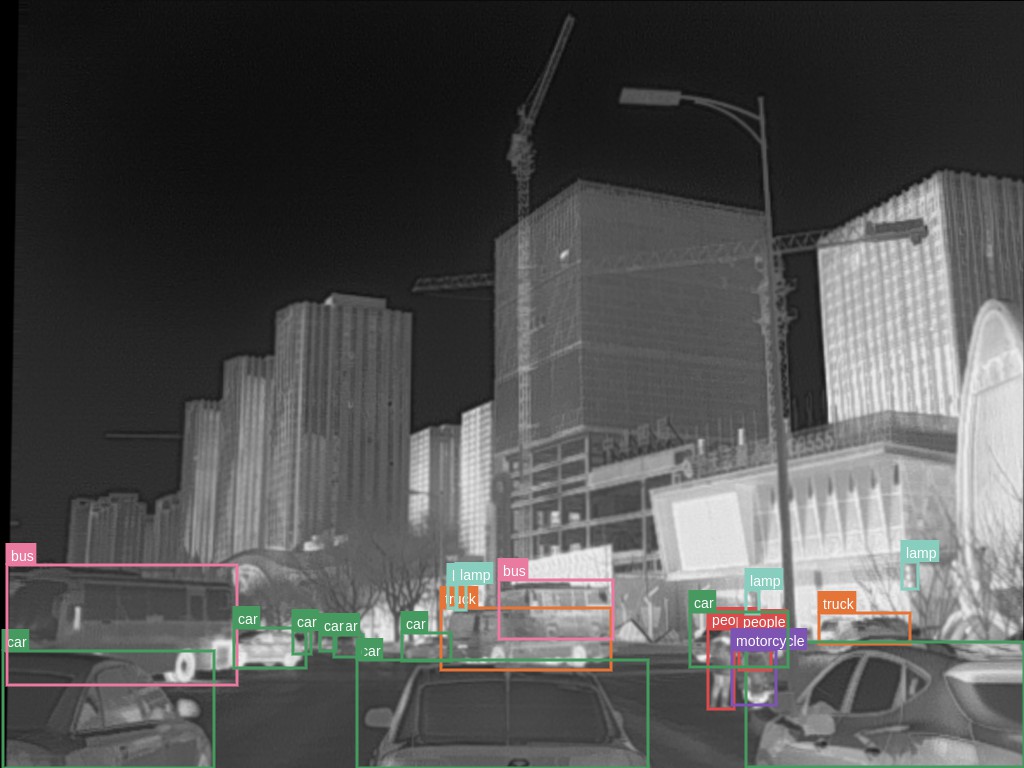}
    \caption{\scriptsize Thermal-GT}
    \end{subfigure}
    \hspace{-1.5mm}
    \begin{subfigure}{0.245\columnwidth}
    \includegraphics[width=1\columnwidth]{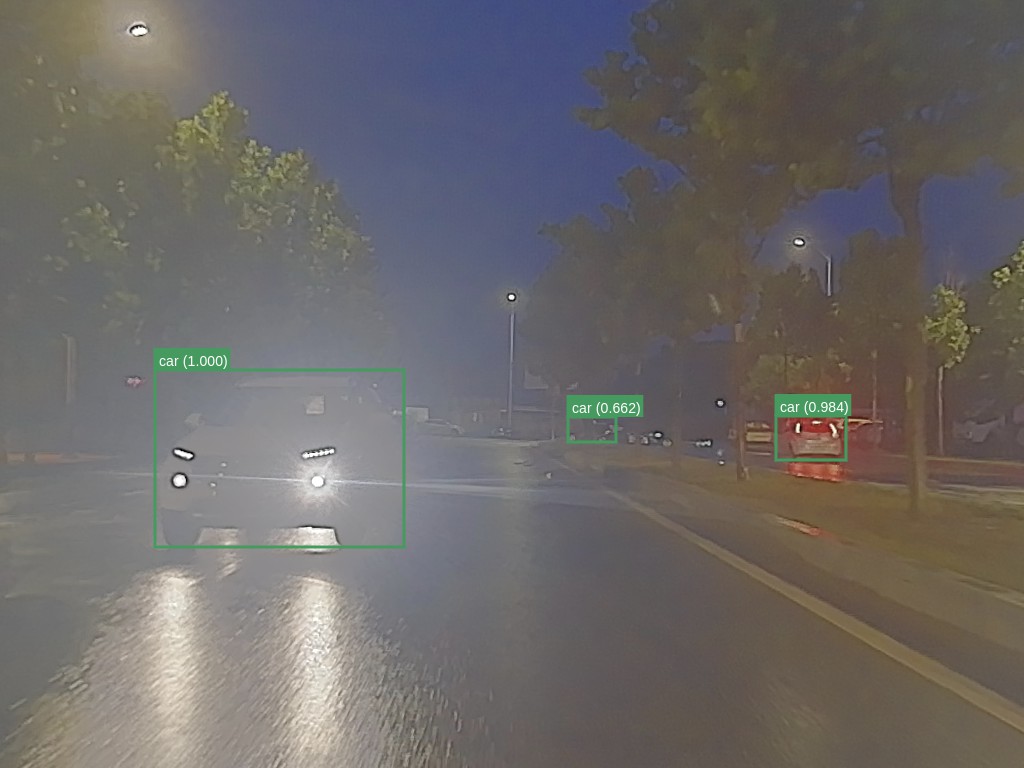} 
    \includegraphics[width=1\columnwidth]{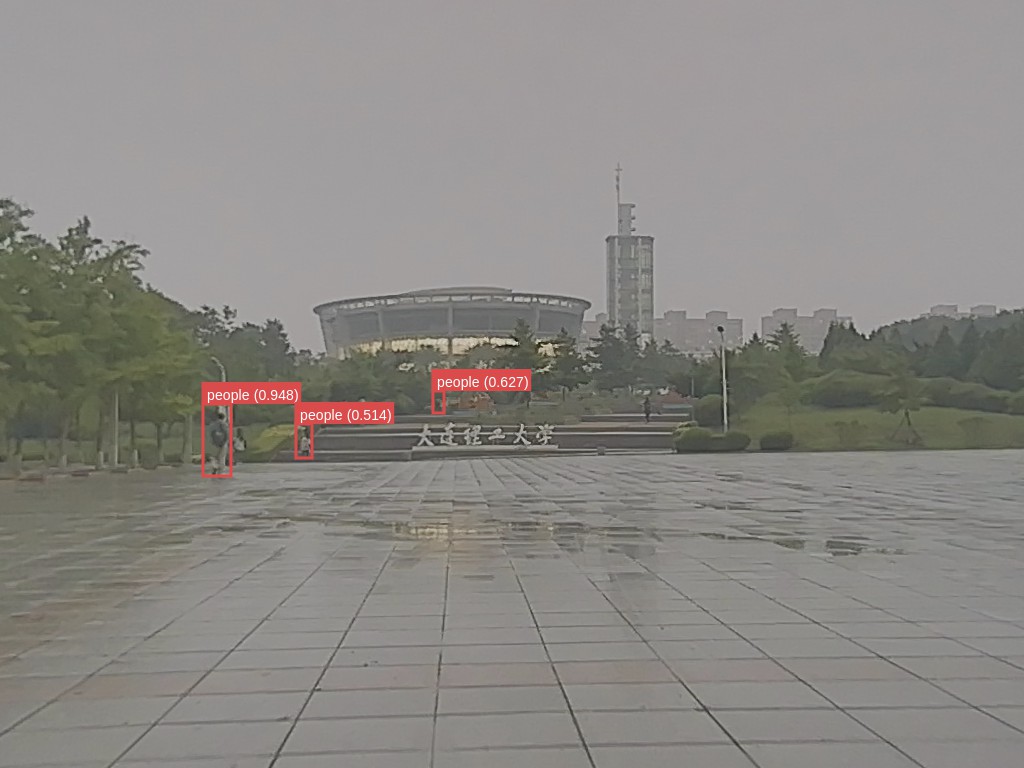}
    \includegraphics[width=1\columnwidth]{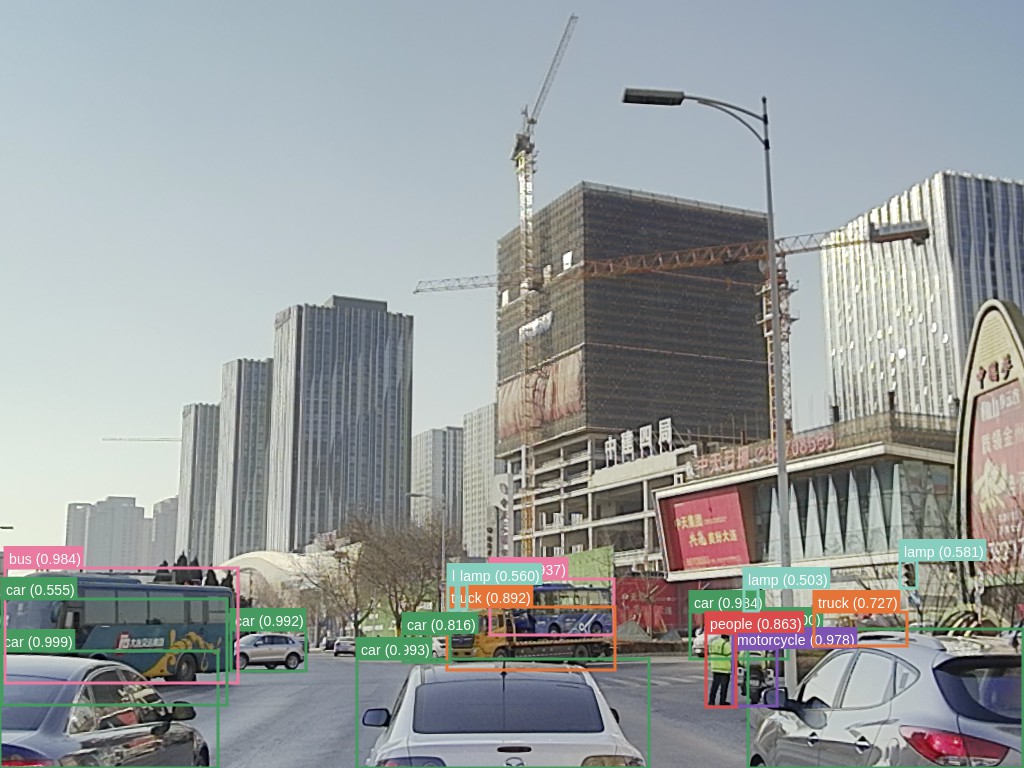}
    \caption{\scriptsize Agnostic CBAM}
    \end{subfigure}
    \hspace{-1.5mm}
    \begin{subfigure}{0.245\columnwidth}
    \includegraphics[width=1\columnwidth]{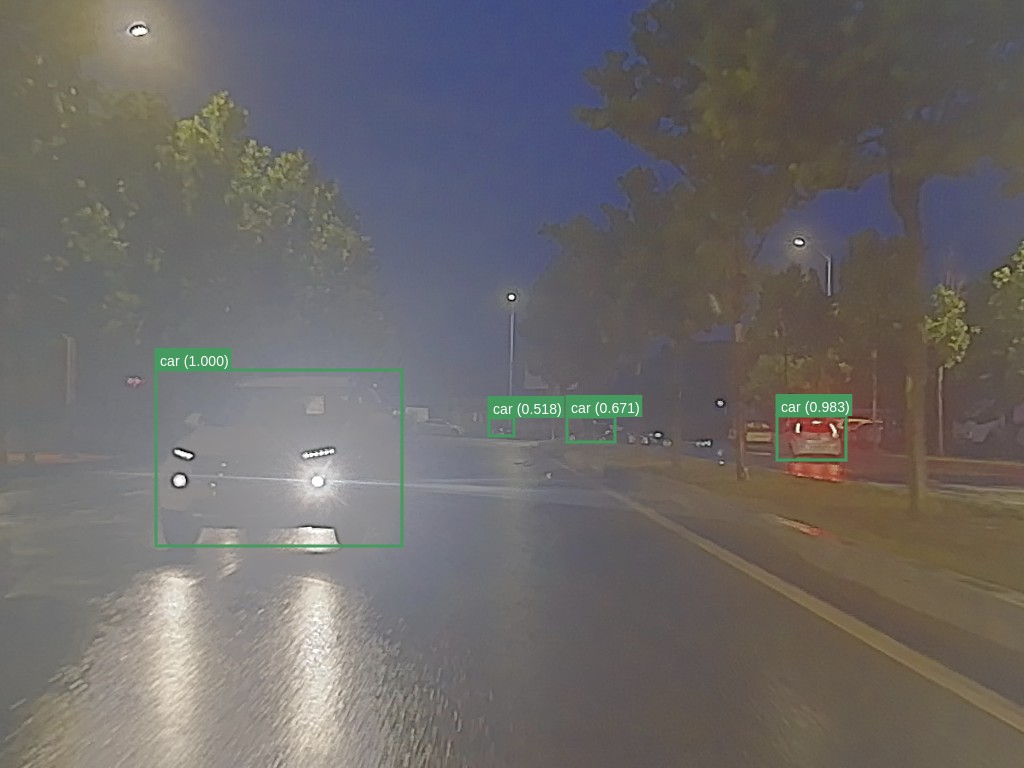} 
    \includegraphics[width=1\columnwidth]{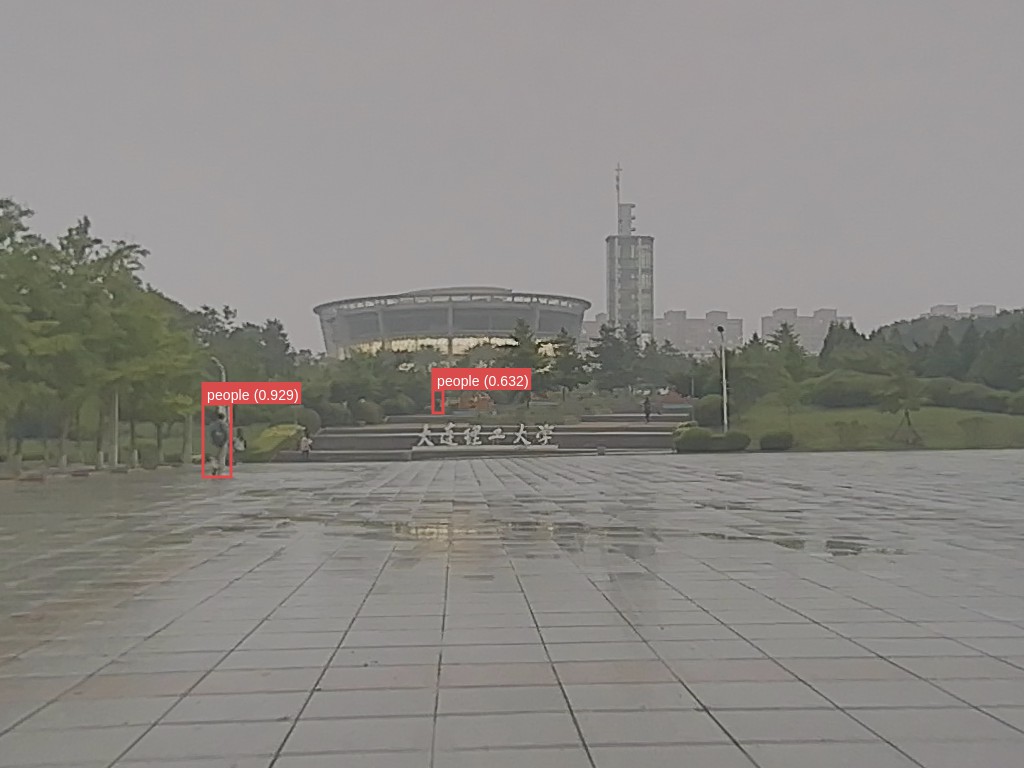}
    \includegraphics[width=1\columnwidth]{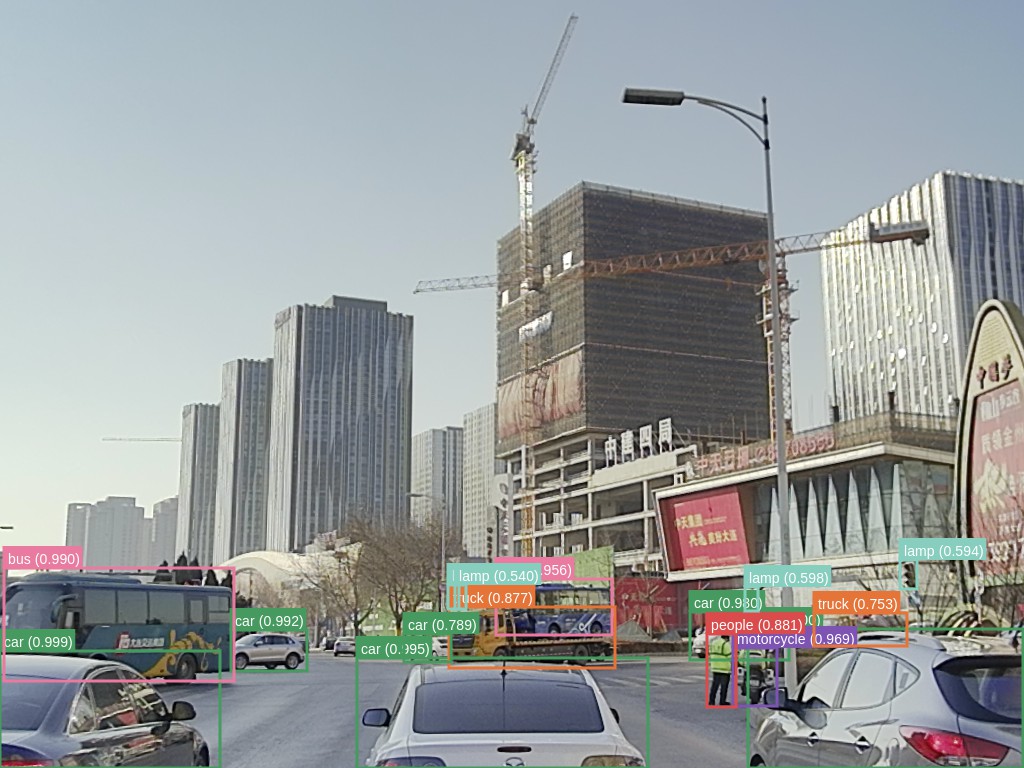}
    \caption{\scriptsize Adaptive CBAM}
    \end{subfigure}
    \caption{Example of failure cases on M$^3$FD dataset. Both models struggled with distant small objects in \emph{night} and \emph{overcast} images and cluttered objects in \emph{day} images.}
    \label{fig:results-failure}
\end{figure}

\textbf{Scene-Adaptive Object Detection:}
This subsection reports quantitative and qualitative object detection results of our proposed methods, compared with existing works. From~\cref{tab:result-m3fd}, our scene-adaptive CBAM model outperforms existing methods on the M$^3$FD dataset using the mean Average Precision \mbox{$\text{IoU}=0.5$} (mAP@0.5) metric used in~\cite{liang2023explicit, liu2022target}. On the \emph{full} test set, it outperforms EAEFNet \cite{liang2023explicit} by $1.4\%$ and the scene-agnostic CBAM model (in which only one set of CBAM fusion modules are trained using all training images) by $1\%$. A comparison of qualitative detection results on M$^3$FD dataset between the scene-agnostic and scene-adaptive models is shown in~\cref{fig:results-m3fd}. From the zoomed-in area of the figures, we can see that the scene-adaptive model detects some occluded, blurred objects that the scene-agnostic model fails to detect. Note that, the single-modality models used in this experiment are pretrained on COCO and further fine-tuned on the M$^3$FD training set for better performance. We also show some failure cases on M$^3$FD in~\cref{fig:results-failure} where both fusion models struggled with distant small objects in \emph{overcast} and \emph{night} scenes, and cluttered objects under daylight.

\begin{figure}
    \centering
    \begin{subfigure}{.24\columnwidth}
    \includegraphics[width=1\columnwidth, trim={1.2cm 0.8cm 1.5cm 1cm}, clip]{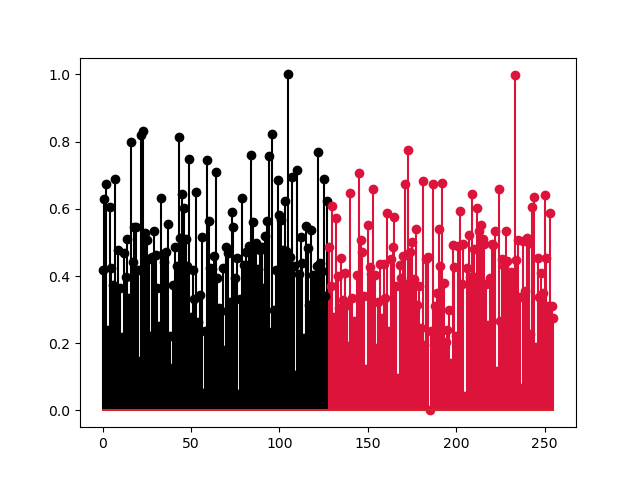}
    \caption{\scriptsize M$^3$FD-Day}
    \end{subfigure}
    \begin{subfigure}{.24\columnwidth}
    \includegraphics[width=1\columnwidth, trim={1.2cm 0.8cm 1.5cm 1cm}, clip]{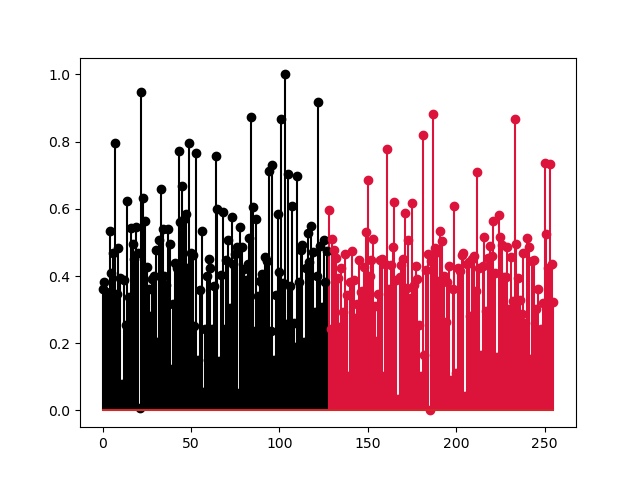}
    \caption{\scriptsize M$^3$FD-Overc.}
    \end{subfigure}
    \begin{subfigure}{.24\columnwidth}
    \includegraphics[width=1\columnwidth, trim={1.2cm 0.8cm 1.5cm 1cm}, clip]{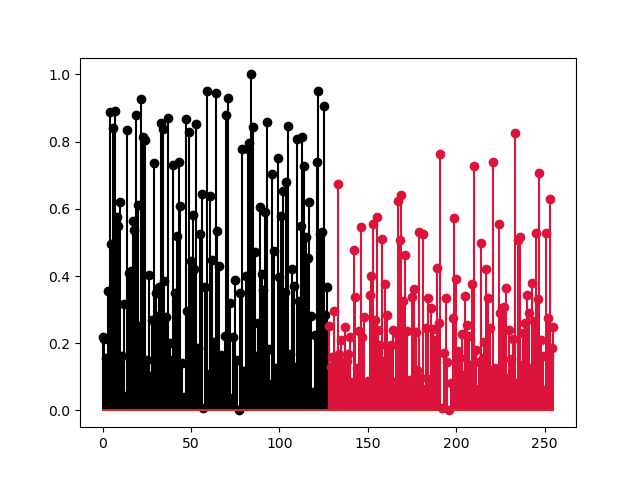}
    \caption{\scriptsize M$^3$FD-Night}
    \end{subfigure}
    \begin{subfigure}{.24\columnwidth}
    \includegraphics[width=1\columnwidth, trim={1.2cm 0.8cm 1.5cm 1cm}, clip]{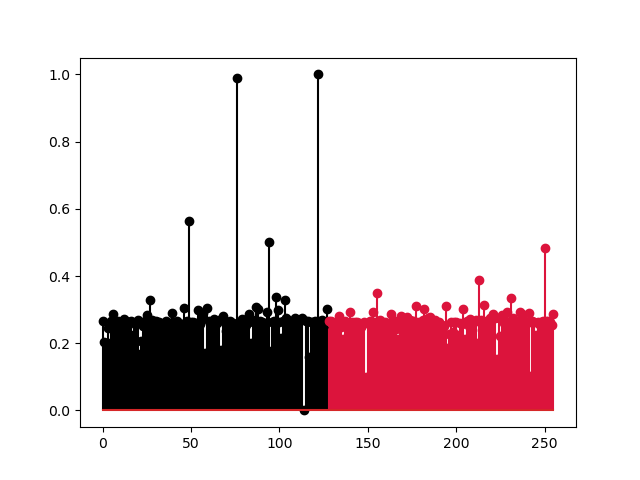}
    \caption{\scriptsize M$^3$FD-Chall.}
    \end{subfigure} \\
    \begin{subfigure}{.24\columnwidth}
    \includegraphics[width=1\columnwidth, trim={1.2cm 0.8cm 1.5cm 1cm}, clip]{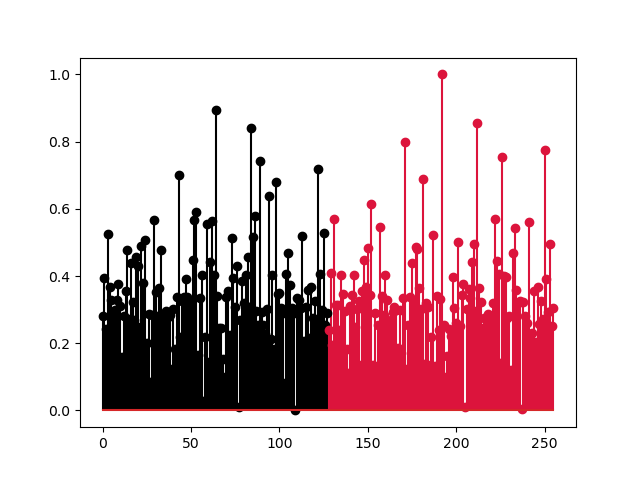}
    \caption{\scriptsize FLIR-Day}
    \end{subfigure}
    \begin{subfigure}{.24\columnwidth}
    \includegraphics[width=1\columnwidth, trim={1.2cm 0.8cm 1.5cm 1cm}, clip]{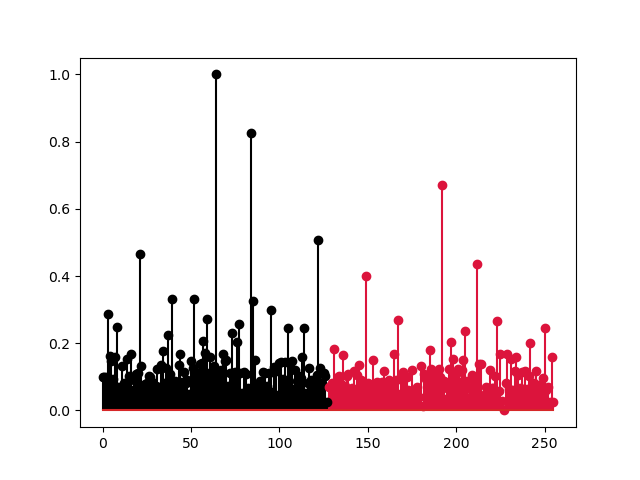}
    \caption{\scriptsize FLIR-Night}
    \end{subfigure}
    \begin{subfigure}{.24\columnwidth}
    \includegraphics[width=1\columnwidth, trim={1.2cm 0.8cm 1.5cm 1cm}, clip]{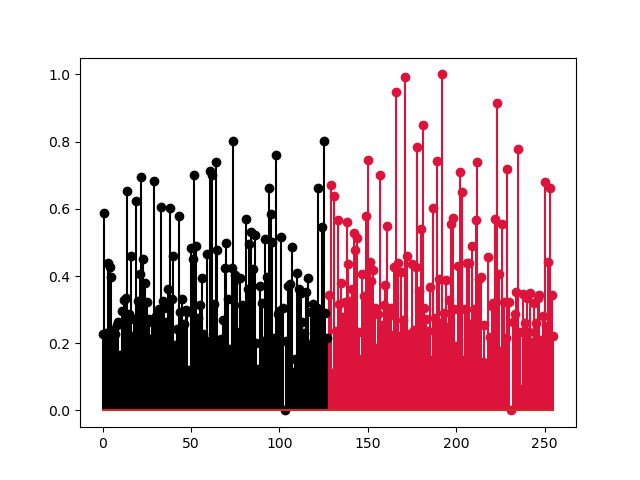}
    \caption{\scriptsize FLIR-Full}
    \end{subfigure} 
    \begin{subfigure}{.24\columnwidth}
    \includegraphics[width=1\columnwidth, trim={1.2cm 0.8cm 1.5cm 1cm}, clip]{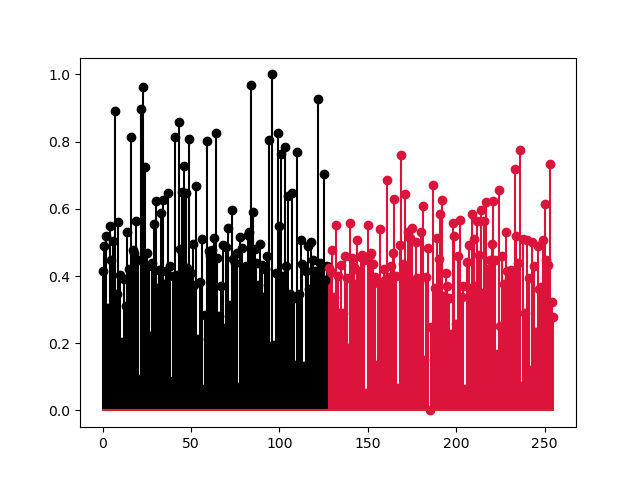}
    \caption{\scriptsize M$^3$FD-Full}
    \end{subfigure} \\
    
    \caption{Normalized attention weights for 256 feature channels in CBAM fusion module trained on different scenes. Thermal channels are in black, and RGB channels in crimson. The fusion module trained on the entire dataset (g-h) exhibits similar attention patterns across all scene/weather conditions, whereas from \emph{day} to \emph{overcast} to \emph{night}, the scene-specific fusion module (a-f) attends increasingly on thermal features.}
    \label{fig:results-visual-cbam}
\end{figure}

\begin{table*}
    \centering
    \caption{Quantitative detection AP on the \emph{clear} scene and unseen scenes for \emph{car} following the KITTI evaluation~\cite{geiger2012we} used in~\cite{Bijelic_2020_STF}. Models are all trained on the training set of the \emph{clear} scene. Our scene-adaptive CBAM model is trained on \emph{clear-day} and \emph{clear-night} splits.}
    \label{tab:stf-clear}
    \resizebox{0.98\textwidth}{!}{
        \begin{tabular}{l|ccc|ccc|ccc|ccc}
        \hlineB{3}
            \multirow{3}{*}{Method} 
            & \multicolumn{12}{c}{Test Scene} \\ \cline{2-13}
            & \multicolumn{3}{c|}{Clear} & \multicolumn{3}{c|}{Light Fog} & \multicolumn{3}{c|}{Dense Fog} & \multicolumn{3}{c}{Snow/Rain} \\ %
            & easy & mod. & hard & easy & mod. & hard & easy & mod. & hard &  easy & mod. & hard \\
            \hlineB{3}
            RGB only   & 90.14 & 87.56 & 80.87 & 91.19 & 88.47 & 82.02 & 90.43 & 85.59 & 80.79 & 89.44 & 82.87 & 77.81 \\
            Gated only & 88.51 & 80.09 & 74.65 & 87.98 & 78.92 & 73.59 & 80.52 & 75.86 & 70.42 & 80.58 & 75.59 & 69.52 \\
            \hline
            Fusion SSD~\cite{Bijelic_2020_STF}  & 87.73 & 78.02 & 69.49 & 88.33 & 78.65 & 76.54 & 74.07 & 68.46 & 63.23 & 85.49 & 75.28 & 67.48 \\
            Deep Fusion~\cite{Bijelic_2020_STF} &  90.07 & 80.31    & 77.82 & 90.60 & 81.08 & 79.63 & 86.77 & 77.28 & 73.93 & 89.25 & 79.09 & 70.51 \\
            Deep Entropy Fusion~\cite{Bijelic_2020_STF} & 89.84 & 85.57 & 79.46 & 90.54 & 87.99 & 84.90 & 87.68 & 81.49 & 76.69 & 88.99 & 83.71 & 77.85 \\
            Scene-agnostic CBAM \textbf{(ours)}    
            & \textbf{90.33} & 88.53 & \textbf{81.16} & \textbf{91.43} & 89.05 & \textbf{84.94} & 90.75 & \textbf{88.66} & \textbf{82.07}& \textbf{89.99} & \textbf{86.57} & \textbf{79.79} \\
            Scene-adaptive CBAM \textbf{(ours)}    
            & 90.29 & \textbf{88.53} & 81.07 & 91.13 & \textbf{89.13} & 84.20 & \textbf{90.77} & 88.37 & 81.68 & 89.96 & 86.30 & 79.74 \\
            \hlineB{3}
        \end{tabular}}
\end{table*}

\begin{table*}
    \centering
    \caption{Quantitative detection AP on all scenes for \emph{pedestrian}, \emph{truck}, \emph{car}, and \emph{cyclist} following the KITTI evaluation~\cite{geiger2012we} used in~\cite{Bijelic_2020_STF}. Models are trained on the training set of all scenes. The last column shows mAP@0.5 for all objects on all test images.}
    \label{tab:stf-full}
    \resizebox{0.98\textwidth}{!}{
        \begin{tabular}{l|ccc|ccc|ccc|ccc|cccc}
        \hlineB{3}
            \multirow{3}{*}{Method} 
            & \multicolumn{16}{c}{Test Scene} \\ 
            \cline{2-17}
            & \multicolumn{3}{c|}{Clear} & \multicolumn{3}{c|}{Light Fog} & \multicolumn{3}{c|}{Dense Fog} & \multicolumn{3}{c|}{Snow/Rain} & \multicolumn{4}{c}{Full} \\ %
            & easy & mod. & hard & easy & mod. & hard & easy & mod. & hard &  easy & mod. & hard & easy & mod. & hard & all \\
            \hlineB{3}
            RGB only 
            & 87.05 & 83.88 & 82.93 & 89.68 & 88.88 & 87.99 & 88.61 & 88.28 & 87.90 & 88.92 & 86.01 & 83.73 & 84.22 & 79.94 & 76.30 & 80.85 \\
            Gated only & 
            81.69 & 76.19 & 74.57 & 85.63 & 84.01 & 80.19 & 83.40 & 82.00 & 79.88 & 84.03 & 79.54 & 77.38 & 80.70 & 73.58 & 70.13 & 75.15 \\
            \hline
            Scene-agnostic CBAM \textbf{(ours)}    
            & \textbf{88.65} & 85.12 & \textbf{84.25} & 90.30 & \textbf{89.68} & \textbf{88.95} & 89.78 & 89.18 & 88.82 & 89.25 & 87.01 &  \textbf{85.77} & 86.11 & 81.84 & \textbf{78.52} & 83.01 \\
            Scene-adaptive CBAM \textbf{(ours)}    
            & 88.60 & \textbf{85.24} & 84.22 & \textbf{90.53} & 89.39 & 88.89 & \textbf{89.79} & \textbf{89.33} & \textbf{89.03} & \textbf{89.37} & \textbf{87.46} & 85.69 & \textbf{86.13} & \textbf{81.85} & 78.48 & \textbf{83.11} \\
            \hlineB{3}
        \end{tabular}}
\end{table*}

For the FLIR Aligned dataset, we evaluate fusion networks built from an RGB network pretrained on COCO and a thermal network trained on the unaligned FLIR thermal training set. In general, both our scene-agnostic and scene-adaptive fusion models outperform the baselines by a large margin (\cref{tab:flir-results}), due to the increase in data the thermal and RGB networks had access to. Some qualitative detection results on FLIR test images along with attention visualizations are given in \cref{fig:results-flir}. We observe that scene-adaptive model tends to detect bicycles more successfully than scene-agnostic model, especially when the bicycle is rode by a person (see row 2 and 6 in~\cref{fig:results-flir}). The higher margin of AP@0.5 for \emph{bicycle} in~\cref{tab:flir-results} also aligns with this observation. In order to exam the effects of scene-adaptive CBAM, we visualize the CBAM using class activation map (CAM)~\cite{muhammad2020eigen} where the spatial attention is shown by a heat map. From the visualization, we can see there is generally no difference between scene-agnostic CBAM and scene-adaptive CBAM for day images. However, the spatial attention in scene-adaptive CBAM attend more on small areas.

We visualize the channel attention of the scene-specific fusion module by plotting the normalized attention weights of thermal (black) and RGB (crimson) features for various scenes in~\cref{fig:results-visual-cbam}. Higher value implies CBAM attends more on that feature channel. We find that scene-agnostic CBAM exhibits similar channel attention patterns across all scenes, while scene-adaptive CBAM shows tailored attention patterns per scene. Moreover, we observe attention weight increases on thermal features compared with RGB features from day to overcast to night images, likely as RGB images contain less information under lower illumination.

For the STF dataset, we first follow~\cite{Bijelic_2020_STF} and train our fusion modules only on \emph{clear-day} and \emph{clear-night} RGB-gated image pairs for fair comparison. As shown in \cref{tab:stf-clear}, the scene-agnostic and scene-adaptive CBAM models achieve similar performance on different scenes and outperform the baseline models using even more modalities than RGB-gated images~\cite{Bijelic_2020_STF}. When training on all scenes in \cref{tab:stf-full}, we can see that our scene-adaptive model outperforms the scene-agnostic model by $0.1\%$ on mAP@0.5. Single-modality models used for this experiment are also further trained on STF training data, due to their use of 10 and 12 bit gated and RGB imagery. \cref{fig:results-visual-stf} presents a few examples of the qualitative detection results in various scenes.

\begin{table}
    \caption{Ablation study on different fusion modules. Object detection results (mAP@0.5) on M$^3$FD dataset are reported.}
    \label{tab:ablation-m3fd}
    \centering
    \resizebox{\columnwidth}{!}{
        \begin{tabular}{l|c|cccccc|}
            \hlineB{3}
            \multirow{2}{*}{Fusion Module} & \multirow{2}{*}{\shortstack[c]{Train/Search \\Scene}} & \multicolumn{5}{c}{Test Scene} \\ \cline{3-7}
            & & Day & Night & Overcast & Challenge & Full \\
            \hlineB{3}
            RGB only        &  \multirow{11}{*}{Full}   & 71.59         & 91.06           & 81.55             & 80.03              & 77.79         \\
            Thermal only    &       & 65.68         & 89.17           & 79.66             & 76.39              & 74.63         \\
            ECAAttn (Tr)        &       & 73.38        & 93.39           & 83.55             & 82.28              & 80.17        \\
            ECAAttn (RH)     &       & 72.25         & 92.83           & 81.98             & 80.53             & 78.81         \\
            ECAAttn (TH)             &       & 74.02         & 93.38           & \bf 84.25             & \bf 81.48              & 80.32         \\
            ShuffleAttn (Tr)    &       & 73.47         & \bf 94.56           & 84.61             & 80.91              & 80.17      \\
            ShuffleAttn (RH) &       & 72.78        & 92.63           & 83.61            & 80.37             & 79.28         \\
            ShuffleAttn (TH)             &       & 74.07         & 93.21           & 84.19            & 81.43              & 80.34        \\
            
            CBAM (Tr)           &       & 73.11         & 93.01           & 83.11           & 80.17              & 79.33       \\
            CBAM (RH)        &       & 72.85         & 92.46           & 83.54             & 80.73            & 79.21         \\
            CBAM (TH)             &       & \bf 74.53         & 93.09           & 84.11             & 81.06             & \bf 80.46 
               \\
            \hline
            \multirow{4}{*}{\shortstack[l]{ECAAttn \\(TH)}}   
            & Day       & \bf 74.75         & \textbf{94.51 }          & 84.16             & 81.09              & \textbf{80.65}         \\
            & Night     & 72.00         & 91.84           & 83.56             & 79.74             & 78.74         \\
            & Overcast  & 71.96         & 92.67           & \textbf{84.44 }            & 80.09              & 79.18         \\
            & Challenge & 73.25         & 93.11           & 83.78             & \textbf{81.88 }             & 80.14         \\
            \hline                                                        
            \multirow{4}{*}{\shortstack[l]{ShuffleAttn \\(TH)}}   
            & Day       & \textbf{75.28}         & \textbf{94.64}           & \textbf{84.72}             & \textbf{81.85}              & \textbf{81.04}         \\
            & Night     & 71.79         & 92.21           & 83.30             & 78.95              & 78.21         \\
            & Overcast  & 73.57         & 92.42           & 84.32             & 80.95              & 80.00         \\
            & Challenge & 72.42         & 92.90           & 84.21             & 81.27              & 79.62         \\
            \hline
            \multirow{4}{*}{\shortstack[l]{CBAM \\(TH)}}      
            & Day       & \textbf{76.04}         & 94.07          & 84.89            & 80.78              & \textbf{81.07}         \\
            & Night     & 72.68         & 92.55           & 83.20             & 78.77              & 78.62         \\
            & Overcast  & 73.30         & 92.53           & \textbf{85.15}             & 80.67              & 79.94         \\
            & Challenge & 74.10         & \textbf{94.28}           & 82.70             & \textbf{82.61}              & 80.93         \\
            \hline                                                          
            \multirow{5}{*}{DSF-NAS}                                                             
            & Day       & \textbf{75.68}         & \textbf{94.25}           & \textbf{84.35}             & \textbf{81.85}              & \textbf{81.03}         \\
            & Night     & 72.32         & 91.94           & 83.85             & 80.51              & 79.12         \\
            & Overcast  & 73.15         & 93.46           & 83.79             & 80.60              & 79.52         \\
            & Challenge & 72.90         & 93.44          & 83.29             & 81.59              & 79.81         \\
            & Full      & 74.68         & 92.65           & 83.90             & 81.67              & 80.56         \\
            \hlineB{3}
        \end{tabular}
    }
    \captionsetup{font=footnotesize}
    \caption*{Tr -- Trained head \hfill TH -- Thermal head \hfill RH -- RGB head}
\end{table}

\textbf{Computational Benchmarks:} We compiled our CBAM fusion models using TorchInductor and conducted benchmarks on a Titan RTX. The inference time for the scene-adaptive fusion model is 0.032 seconds per individual image pair, while the scene-agnostic variant clocks in at 0.028 seconds. These times are comparable with other recent multimodal object detection approaches (\cref{tab:result-m3fd},~\ref{tab:flir-results}) and meet the speed requirements for real-time autonomous driving applications.

\subsection{Ablation Studies}
\textbf{Fusion Module Design:} We conduct an ablation study using the M$^3$FD dataset to explore the effects of different fusion modules and architectures (\cref{tab:ablation-m3fd}). We compare our CBAM-based RGB-X fusion approach against two other attention modules: ECAAttn~\cite{wang2020eca, deevi2022expeditious} and ShuffleAttn~\cite{zhang2021sa}. Furthermore, we also compare against custom fusion modules (DSF-NAS) designed purposely for this fusion task via neural architecture search. In particular, we use Bilevel Multimodal Neural Architecture Search~\cite{yin2022bm} (BM-NAS) to automate this design as its gradient-based optimization approach makes it faster compared to other NAS methods based on reinforcement learning and genetic algorithms. Specifically, we allow BM-NAS to optimize over sequential applications of two operations chosen from sum, spatial attention, channel attentions from CBAM and ECAAttn, and 2D convolution of concatenated features.

We first train for fusion using scene-agnostic CBAM, ECAAttn, and ShuffleAttn modules along with either a trainable, frozen thermal, or frozen RGB detector head. We find that training with a frozen detector head initialized with thermal weights performs the best in~\cref{tab:ablation-m3fd}, possibly due to the lower variance of thermal data across different scenes. We repeat the study under the scene-adaptive regime, with the previous three attention modules and frozen thermal detection heads, along with DSF-NAS fusion modules. Overall, we find similar performance between DSF-NAS and CBAM-based fusion networks. However, CBAM fusion models exhibit better performance on scene-specific data verifying its use in our proposed modular framework.

\begin{table}
\centering
\caption{Object detection results (mAP@0.5) of our scene-adaptive CBAM model trained using decreasing amounts of data.}
    \label{tab:dataset-size}
\resizebox{0.8\linewidth}{!}{%
\begin{tabular}{c|c|c|c|c} 
\hlineB{3}
\multirow{2}{*}{Dataset}            & \multicolumn{4}{c}{\% of Original Training Set} \\ \cline{2-5} 
               & 100\% & 50\%    & 25\% & 1\%  \\ 
\hlineB{3}
FLIR               & 86.16                  & 85.70                    & 84.60        &  75.72        \\
M$^3$FD               & 81.46                  & 78.34                    & 77.65            &  41.94         \\
STF-Clear & 80.65                  & 80.73                    & 80.10                &    73.76      \\
STF-Full & 83.11                  & 83.06                    & 82.99                &   75.64       \\
\hlineB{3}
\end{tabular}
}
\end{table}

\textbf{Effect of Training Dataset Size on Fusion: }
As our proposed fusion method looks to fuse pretrained networks with lightweight fusion modules, the fusion process should still be effective and be able to generalize even when done with limited amounts of training data. To determine the extent of this, we perform fusion using 100\%, 50\%, 25\%, and 1\% of the original datasets in~\cref{tab:dataset-size}. Overall, we find that competitive results can still be achieved using only 25\% of the original training data results, with the exception of M$^3$FD which decays quicker than the rest.

\begin{table}
\centering
\caption{Object detection results (mAP@0.5) of our scene-adaptive CBAM model on unknown scenes in M$^3$FD dataset.}
\label{tab:classifier-exp}
\resizebox{0.8\columnwidth}{!}{
\begin{tabular}{c|cccc}
\hlineB{3}
    \multirow{2}{*}{Test} & \multicolumn{4}{|c}{Excluded Training Scene} \\ \cline{2-5}
    & Day & Night & Overcast & Challenge \\
    \hline
    Excluded Scene & 73.17 & 93.50 & 84.18 & 80.74 \\
    All Scenes & 80.48 & 81.30 & 81.27 & 80.98 \\
\hlineB{3}
\end{tabular}}
\end{table}

\textbf{Performance on Unknown Scenes:} As our proposed method requires scenes to be known during training, we further investigate the performance of our method on unknown/unexpected scenes. In this experiment, our scene-specific CBAM fusion modules and scene classifiers are trained with one scene data excluded, and tested on that excluded scene and all test images. We observed minor regression in overall performance (row 2 in~\cref{tab:classifier-exp}) compared with our scene-adaptive model trained on all scenes (81.46 in~\cref{tab:result-m3fd}), which is expected as there is no fusion module trained specifically for that unknown scene. However, the overall mAP@0.5 in all cases is still higher than scene-agnostic model trained on all scenes (80.46 in~\cref{tab:result-m3fd}). Specifically, in the case of \emph{night} or \emph{overcast} scene excluded, the object detection performance on the unknown scene (row 1 in~\cref{tab:classifier-exp}) is higher than the scene-agnostic model. This is possibly because our scene classifier tends to select a fusion module trained on a similar scene, for instance, classifying \emph{night} image as \emph{overcast} and vice versa.

\section{Limitations and Future Work}
\label{sec:limitations}
Our method requires aligned RGB-X data, which is not always available. The scene-specific modules require scenes to be known during training, and the approach is not expected to work as well in unexpected weather conditions. Future work looks to incorporate unsupervised~\cite{gan2023unsupervised} and online learning~\cite{lee2023online} to adapt to unexpected conditions.

\section{Conclusion}
\label{sec:conclusion}

We presented a novel RGB-X object detection model that improves autonomous vehicle perception in different weather and lighting conditions. We showed that our method is superior compared to existing works on two RGB-T and one RGB-gated object detection benchmarks, demonstrating the robustness of our scene-adaptive models and generalizability to different modalities. Furthermore, our use of lightweight fusion modules brings us closer to achieving a more modular design for deep sensor fusion. For future work, we look to train and leverage larger pretrained models for both RGB and thermal modalities via multitask learning and to incorporate into an online learning framework to adapt to unexpected weather patterns.

{\small
\bibliographystyle{ieee_fullname}
\bibliography{egbib}
}

\end{document}